%% file: main.tex
\documentclass[lettersize,journal]{IEEEtran}
\usepackage{amsmath,amsfonts}
\usepackage{algorithmic}
\usepackage{array}
\usepackage[caption=false,font=normalsize,labelfont=sf,textfont=sf]{subfig}
\usepackage{textcomp}
\usepackage{stfloats}
\usepackage{url}
\usepackage{verbatim}
\usepackage{graphicx}
\usepackage{booktabs}
\usepackage{marvosym}
\usepackage{multirow}
\def\BibTeX{{\rm B\kern-.05em{\sc i\kern-.025em b}\kern-.08em
    T\kern-.1667em\lower.7ex\hbox{E}\kern-.125emX}}
\usepackage{balance}
\begin{document}
\title{SuperUDF: Self-supervised UDF Estimation for Surface Reconstruction}
\author{Hui Tian\textsuperscript{1}, Chenyang Zhu\textsuperscript{1 \Letter},
Yifei Shi\textsuperscript{1},
Kai Xu\textsuperscript{1 \Letter}
\thanks{1, National University of Defense Technology}}

\markboth{Journal of \LaTeX\ Class Files,~Vol.~18, No.~9, September~2020}%
{}

\maketitle


\begin{abstract}
Learning-based surface reconstruction based on unsigned distance functions (UDF) has many advantages such as handling open surfaces. We propose SuperUDF, a self-supervised UDF learning which exploits a learned geometry prior for efficient training and a novel regularization for robustness to sparse sampling. The core idea of SuperUDF draws inspiration from the classical surface approximation
operator of locally optimal projection (LOP). The key insight is that if the UDF is estimated correctly, the 3D points should be locally projected onto the
underlying surface following the gradient of the UDF. Based on that, a number of inductive biases on UDF geometry and a pre-learned geometry prior are devised to learn UDF estimation efficiently. A novel regularization loss is proposed to make SuperUDF robust to sparse sampling. Furthermore, we also contribute a learning-based mesh extraction from the estimated UDFs. Extensive evaluations demonstrate that SuperUDF outperforms the state of the arts on several public datasets in terms of both quality and efficiency. Code url is https://github.com/THHHomas/SuperUDF.
\end{abstract}
 

\input{1intro.tex}

\input{2related.tex}
\input{3method.tex}
\input{4exp.tex}

\input{5conclusion.tex}

\clearpage
\bibliographystyle{IEEETran}
\bibliography{bibfile}
\clearpage
\input{6appendix}

\end{document}

%% file: 1intro.tex
\section{Introduction}
\label{intro}

Surface reconstruction from 3D point clouds has been a long-standing problem in graphics and vision. Since the seminal work of Poisson surface reconstruction~\cite{2013Poisson}, there have been a large body of literature~\cite{berger2017survey}. Albeit relying on normal as input, the idea of reconstruction based on implicit field has been inspiring many deep learning methods~\cite{2019DeepSDF, 2020Convolutional,2019Occupancy, 2020Points2Surf, 2020npull}.
Signed Distance Function (SDF) is a typical implicit representation of 3D shapes of arbitrary geometry and topology. Many deep learning models have been proposed to predict the SDF of a 3D point cloud for high-quality reconstruction~\cite{2019DeepSDF, 2020Convolutional, 2020Points2Surf, 2022POCO}. However, a major drawback of SDF is that it can represent only closed and watertight surfaces due to its nature of inside/outside discrimination, so SDF-based methods find difficulty in handling open surfaces such as garments or incomplete scans.

Unsigned Distance Function (UDF) is suited for representing open surfaces since it does not differentiate between inside and outside. Compared to SDF, UDF is easier to learn since it concerns only distance information and ignores the sign. The pioneering work of~\cite{2020ndf} proposes a direct rendering method based on UDFs. However, both this work and its follow-ups~\cite{venkatesh2021deep,zhao2021learning} require strong supervision of ground-truth UDFs. Moreover, surface extraction from UDFs is difficult due to the absence of signs which is critical to marching cube~\cite{lorensen1987marching}.~\cite{2021MeshUDF} convert UDFs to meshes via elevating a UDF to an SDF which brings back the limitation of SDFs.

\input{figures/teaser.tex}

Recently,~\cite{capudf} introduces a self-supervised pipeline to learn smooth UDFs directly from raw point clouds. Their network is trained in a shape-specific manner, requiring $\sim$20 minutes for overfitting one shape. Furthermore, the method is sensitive to the sampling density of the input point cloud; the prediction of field gradients is inaccurate for sparsely sampled point clouds. Inspired by recent works~\cite{2019Occupancy,2020Convolutional} demonstrating that geometry priors learned from large datasets benefit efficient reconstruction, we propose SuperUDF, a self-supervised learning method for fast UDF estimation with learned priors. To obtain robustness under sparse sampling, we introduce a novel regularization loss to train SuperUDF. We also introduce a learning-based mesh extraction directly from the estimated UDFs.

The core idea of our self-supervised UDF estimation draws inspiration from the seminal surface approximation operator of Locally Optimal Projection (LOP)~\cite{lop}. In particular, we first upsample the input point cloud via per-point duplication and random perturbation. The key insight behind our design is that \emph{if the UDF is estimated correctly, the upsampled points should be locally projected onto the underlying (ground-truth) surface following the gradient of the UDF}. Since the ground-truth surface is unknown \emph{a priori}, we instead require the projected points to approximate the input point cloud.

To do so, we impose a direct constraint on the estimated UDF for a better data approximation by minimizing the mean UDF value of all points of the input point cloud. However, imposing the approximation constraints alone may lead to noisy UDFs where the monotonicity of UDF at one side of the zero-level-set may be violated. To this end, we devise a regularization loss to ensure UDF monotonicity. In particular, for each upsampled point, denoted by $p$, we compute its offset point $q$ by moving $p$ along UDF gradient for a distance of half of $p$'s UDF value. Here, $q$ is expected to reside at the same side as $p$ about the surface. We impose that the UDF gradients at the two points are the same and the UDF value at $q$ should be half of that of $p$. This design also makes the points projected more uniformly.

A difficulty in the method, however, is that the computation of UDF gradients in the losses above is time-consuming, making the training intractable. We thus opt for the reverse. We instead estimate a projection vector for each upsampled point. This results in a projection flow as an approximation of the UDF gradients. The projection flow network can also be trained in a self-supervised manner using the constraints above. The resulting projection flow is a strong geometry prior learned from a shape dataset which greatly improves training efficiency. The UDF can be easily obtained from the projection flow based on their duality.

To learn mesh extraction from UDFs, we partition the 3D space into a regular grid and train a 3D CNN for each grid cell to estimate the signs of its 8 corners. Note that we do not care about the absolute signs being positive or negative, but only concern with their relative signs. In practice, we fix the sign of one corner and let the network predict the rest. Since the network is trained for local sign prediction, it is easy to train and generalizes well across different shapes.

Extensive evaluations demonstrate that SuperUDF outperforms the state
of the arts on ShapeNet, MGN and Scannet in terms of both
quality and efficiency. The main contributions of our work are:
\begin{itemize}
    \vspace{-5pt}\item A self-supervised UDF estimation network inspired by the classical surface approximation operator of Locally Optimal Projection (LOP).
    \item A number of inductive biases on UDF geometry and a pre-learned geometry prior for efficient learning.
    \item A novel regularization loss to make SuperUDF robust to sparse sampling.
    \item A learning-based mesh extraction method that generalizes well across shapes.
\end{itemize}

%% file: figures/teaser.tex
\begin{figure} 
\centering
  \includegraphics[width=\columnwidth]{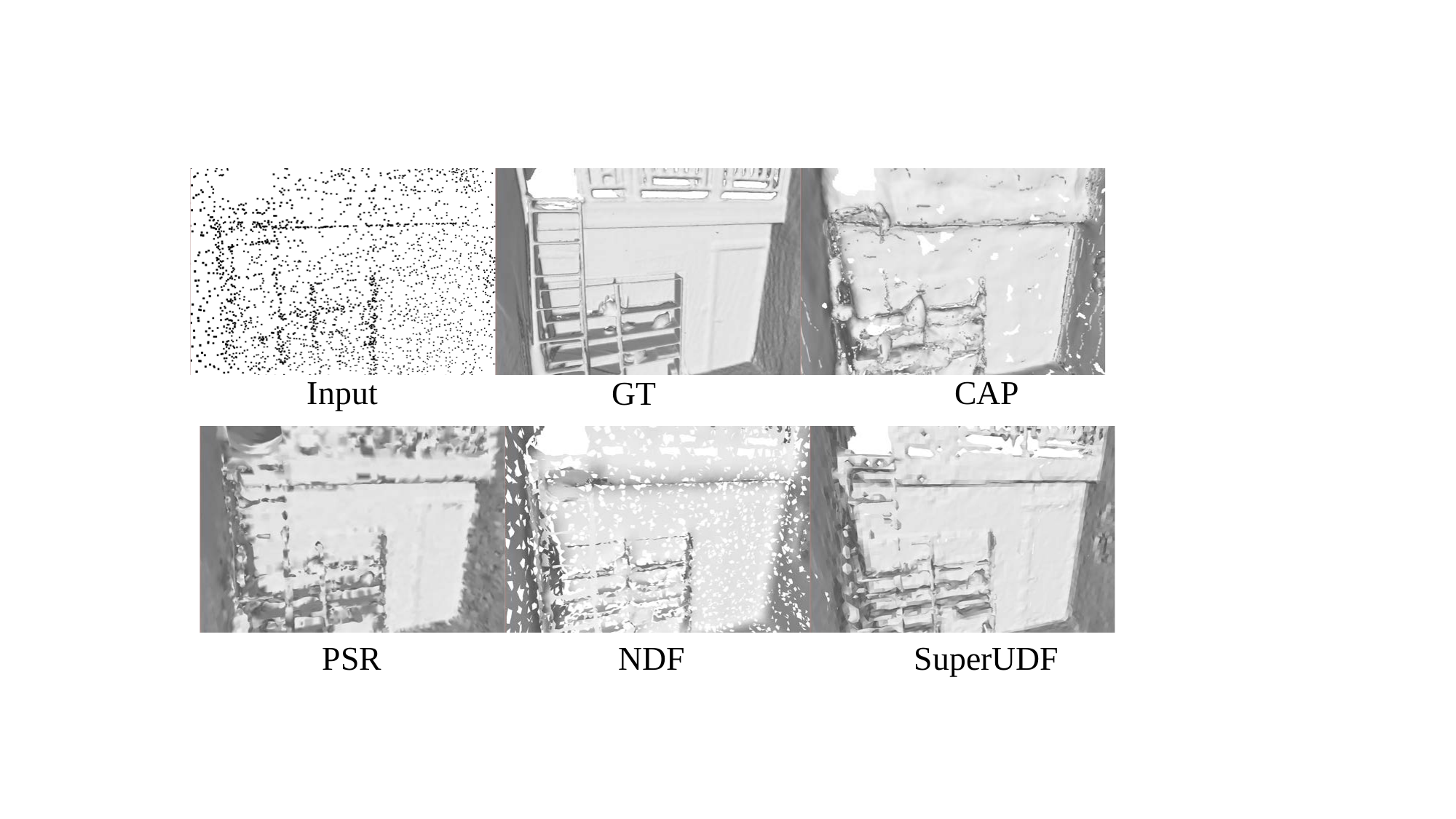}
  \caption{A scene from ScanNet\cite{dai2017scannet} reconstructed with PSR\cite{2013Poisson}, NDF\cite{2020ndf} , CAP\cite{capudf} and our SuperUDF.}
  \label{fig:teaser}
\end{figure}

%% file: 2related.tex
\section{Related Work}

\paragraph{Traditional point cloud surface reconstruction} Point cloud reconstruction has been a long-standing task in graphics and vision. The most important traditional method is Poisson surface Reconstruction~\cite{2013Poisson} and ball-pivoting reconstruction~\cite{ballpivoting}. The former method classifies the query point according to the Poisson indicator function. The latter constructs the continuous surface by making up a ball rolling on the points. Those two methods can reconstruct pretty good surfaces, however, the performance can be improved more.

\paragraph{SDF-based implicit surface reconstruction} 
Deep methods based on SDF always classify the occupancy of query points or directly regress the SDF value, which can be divided into local methods, global methods, and a combination. The global method, as the name implies, when giving a query point, classifies the query according to the whole shape information. The local method classifies the query point according to its neighbor points of it. 
Representative global methods are DeepSDF~\cite{2019DeepSDF}, BSP-Net~\cite{chen2020bsp}. The routine of those methods is extracting the feature code of the whole shape and then recovering the surface from the code. Representative local methods are ConvOccNet~\cite{2020Convolutional}, SSR-Net~\cite{2020SSRNet}, DeepMLS~\cite{2019DeepSDF} and POCO~\cite{2022POCO}. ConvOccNet~\cite{2020Convolutional} first converts the point cloud feature into voxels and then applies volume convolution to enhance the feature of every voxel. SSR-Net~\cite{2020SSRNet} firstly extracts point feature, then maps the neighborhood points feature to octants, and finally classifies the octants. DeepMLS~\cite{deepmls} tries to predict the normal and radius of every point, then classify the query points according to the moving least-squares equation. Furthermore, except for local methods and global methods, there are some methods that try to combine global and local information. The most representative method is Points2Surf~\cite{2020Points2Surf}. It tries to regress the absolute SDF value according to local information and classifies the sign according to global information. 
\input{figures/pipeline.tex}
Another implicit surface reconstruction methods based on SDF are SAL~\cite{SAL}, SALD~\cite{SALD} and On-Surface Prior~\cite{onprior}. This kind of method aims to convert explicit representation, such as point cloud and 3D triangle soup, to implicit SDF representation. Thus, every 3D model needs a unique training process and unique network parameters. SAL~\cite{SAL} uses MLP to predict the SDF of shapes but adopts the UDF-based metric to supervise the network training, SALD ~\cite{SALD} follows the SAL and adds derivative regularization term. On-surface Prior~\cite{onprior} use a pre-trained UDF-based network to help the main network to predict better SDF.

The SDF methods mentioned above achieve excellent progress in point cloud reconstruction. However, SDF representation has its weakness. It is hard to represent an open surface or partial scan. Furthermore, all the deep methods mentioned above needs 3D ground truth as the training label, while 3D ground truth is expensive for closed shape and real scan.

\paragraph{UDF-based implicit surface reconstruction}  
UDF can express more general surfaces, such as open surfaces and partial scans. Lots of works have focus on the UDF representation. NDF~\cite{2020ndf} uses UDF to represent the surface, then they propose a point cloud up-sample method while reconstructing the mesh not directly from UDF but up-sampled point cloud via Ball Pivoting~\cite{ballpivoting} method. DUDE~\cite{dude} represent the shape with a volumetric UDF encoded in a neural network. Those works bring UDF to deep implicit surface reconstruction area and achieve good results. However, they need the 3D ground truth as the training label. What's more, they remain an open problem on how to directly extract iso-surface from UDF. UWED~\cite{uwed} makes use of MLS ( Moving Least Squares ) to convert the UDF to dense point cloud.

\paragraph{Self-supervised method for surface reconstruction} Besides the supervised methods mentioned above, there are a few self-supervised approaches. For example, Neural Pulling~\cite{2020npull} proposes a pipeline to train a network such that the network can predict SDF in the whole space without any extra supervision. However, the method has several disadvantages. First, it needs a dense and complete point cloud as supervision, limiting its scalability in real-world scenarios. Second, the method is only capable of reconstructing objects with closed surfaces. Thus, CAP~\cite{capudf} tries to improve this method by predicting UDF rather than SDF so it can represent open surface. But it also suffers from the requirements of dense point cloud and long inference time. 

\paragraph{Iso-surface extraction} For SDF, the most common method for iso-surface extraction is Marching Cube~\cite{lorensen1987marching}. It is a template matching method based on the sign of eight grid corners. For iso-surface extraction on UDF, there are only a few works. MeshUDF~\cite{2021MeshUDF} uses a neural network to vote the sign of the grid corner while bringing some sign conflict. The two methods try to convert the UDF to SDF. But SDF has difficulty representing the open surface. They are meaningful exploitation of iso-surface extraction on UDF. While sharing the same drawback of SDF and weak generalization ability.

%% file: figures/pipeline.tex
\begin{figure*} 
\centering
  \includegraphics[width=2\columnwidth]{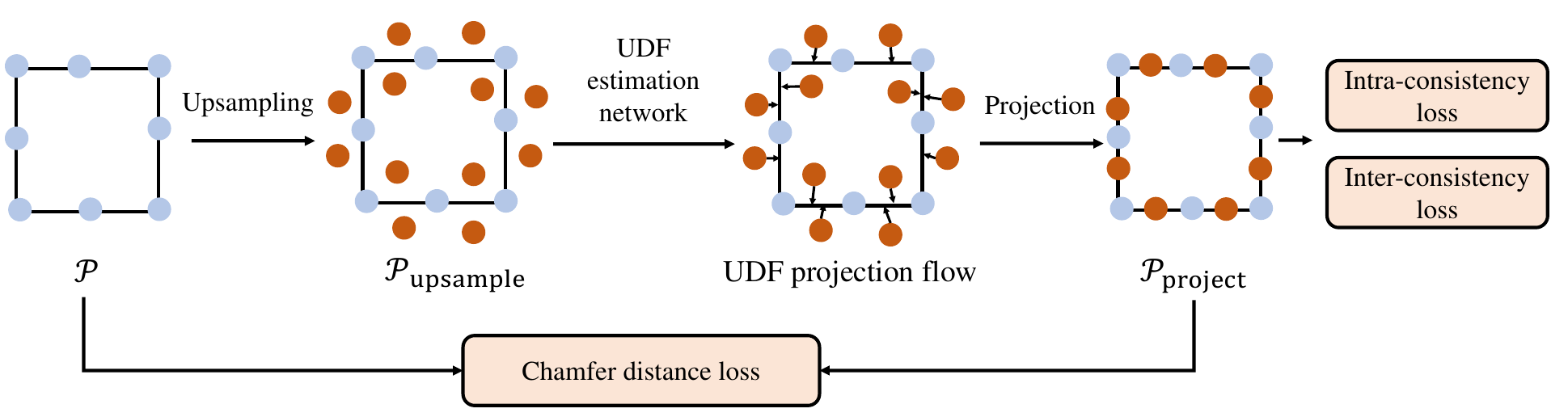}
  \caption{The pipeline of our UDF estimation method. Given a point cloud $\mathcal{P}$, it first performs an upsampling via per-point duplication and perturbation. The upsampled point cloud $\mathcal{P}_\text{upsample}$ is fed into the UDF estimation network. By performing a per-point projection based on the estimated UDF projection flow. The projected points are optimized in a self-supervised manner with several dedicated loss functions.}
  \label{fig:pipeline}
\end{figure*}

%% file: 3method.tex
\section{Method}

\subsection{Overview}\label{sec:overview}
SuperUDF reconstructs the surface by estimating the mesh from a point cloud. 
The whole pipeline contains two main parts: UDF estimation and iso-surface extraction.
In sec.~\ref{sec:udf_est}, we describe the self-supervised UDF estimation network based on the inductive biases on UDF geometry. This part is shown in Fig. \ref{fig:pipeline}
In sec.~\ref{sec:iso_ext}, we introduce a learning-based mesh extraction from the estimated UDFs.
This part is illustrated in Fig. \ref{fig:tiny_net}.

\subsection{Self-supervised UDF Estimation}\label{sec:udf_est}
We propose a method to estimate the UDF from a point cloud $\mathcal{P}$.
As shown in Fig.~\ref{fig:pipeline}, the method starts by upsampling the input point cloud $\mathcal{P}$ via per-point duplication and perturbation, resulting in an upsampled point cloud $\mathcal{P}_\text{upsample}$.
The network then estimates the UDF in an implicit manner: it takes the coordinate of an upsampled point as input and outputs its UDF value, as well as its UDF gradient. The details of the UDF estimation network is illustrated in Fig.~\ref{fig:network}.

\paragraph{Backbone}
Unlike the generic point cloud understanding tasks which require global context, point-wise UDF is supported by local geometry clues.
Directly adopting existing backbones for feature aggregation is sub-optimal.
To alleviate this problem, we develop a local perceptive point transformer to aggregate features from local regions only.
The local perceptive point transformer consists of four Point Transformer layers~\cite{zhao2021pointtrans}, which is shallow and has a perceptive field.
For a point cloud $\mathcal{P}=\{p_i\}^N_1$, we adopt the local perceptive point transformer to extract the point-wise features:
\begin{equation}
    f_i^{(l)} = \sum_{p_j \in \mathcal{N}(p_i)}{\text{PointTransformer}(f_j^{(l-1)}, f_i^{(l-1)})},
\end{equation}
where $\text{PointTransformer}(\cdot)$ is the Point Transformer layer~\cite{zhao2021pointtrans}, $f_i^{(l)}$ is the point feature of $p_i$ at the $l$-th layer, $\mathcal{N}(p_i)$ is the neighboring point set of $p_i$. We compute the neighboring point set by using a KNN with the number of neighboring points being 36.

\input{figures/network.tex}

\paragraph{UDF projection flow estimation}
A straightforward solution to estimate the point-wise UDF is to aggregate features at each upsampled point and output its UDF with a neural network.
For any point, the UDF gradient can then be computed by estimating the dense point-wise UDFs in the neighboring region of that point.
This process is computationally time-consuming, making network training infeasible.
To solve the problem, We propose to directly estimate a projection vector for each upsampled point.
This results in a projection flow as an approximation of the UDF gradients.
The UDF can be obtained from the projection flow based on their duality.
The projection vector which displaces the upsampled point to the surface is estimated in two stages (Fig.~\ref{fig:method_illustration} a). 

First, for each upsampled point $q$, we compute a coarse projection vector by aggregating the estimated approaching vector of the neighboring input points:
\begin{equation}
    s_1 = \frac{1}{k}\sum_{p_x \in \mathcal{N}(q)}{<p_x-q,a_x>a_x}, 
    \label{eq:shift1}
\end{equation}
where $\mathcal{N}(q)$ is the neighboring input points to $q$, $k$ is the number of neighboring points, $<\cdot>$ is the inner-product operation. $a_x$ is an approaching vector of input point $p_x$ which is estimated by:
\begin{equation}
    a_x = \text{MLP}^a(f_x^{(4)}),
\end{equation}
where $\text{MLP}^a(\cdot)$ is the multi-layer perceptron, $f_x^{(4)}$ is the extracted feature of $p_x$. Although we did not provide any direct supervision, we found the learned approaching vector is very similar to the normal vector. Please refer to the supplemental materials for the visualization.
The above process displaces point $q$ such that the displaced position $\hat{q}=q+s_1$ is close to the underlying surface.

\input{figures/method_illustration}

Second, to further optimize the projection flow, we estimate a refining projection vector with a neural network.
To achieve this, we first fetch the features of the neighboring input points $\mathcal{N}(\hat{q})$ at the displaced position $\hat{q}$ and interpolate its feature:
\begin{equation}
    f_{\hat{q}} = \sum_{p_x \in \mathcal{N}(\hat{q})}{[\text{MLP}^f(p_x-\hat{q}) \circ f_x^{(4)}]},
    \label{eq:interpolate}
\end{equation}
where $\text{MLP}^f(\cdot)$ is the multi-layer perceptron, $\circ$ is the element-wise multiplication.
The refining displacement vector can be estimated by:
\begin{align}
    s_2 = \beta\text{Tanh}[\text{MLP}^d(f_{\hat{q}})]
    \label{eq:refine}
\end{align}
where $\text{Tanh}(\cdot)$ is the hyperbolic tangent function, $\text{MLP}^d(\cdot)$ is the multi-layer perceptron, $\beta$ is a scale factor being $0.01$. 
As a result, the projection flow at point $q$ is $s_1+s_2$, and the estimated projection of point $q$ on the underlying scene surface is $\bar{q}=q+s_1+s_2$.

\input{figures/tiny_net.tex}

\paragraph{Self-supervised network training}
If the UDF projection flow is estimated correctly, the upsampled points should be locally projected onto the underlying surface following the flow. Since the underlying surface is unknown a priori, we require the projected points $\mathcal{P}_\text{project}$ to approximate the input point cloud $\mathcal{P}$ instead. A chamfer distance loss between $\mathcal{P}$ and $\mathcal{P}_\text{project}$ is utilized:
\begin{equation}
    L_{\text{chamfer}}=\text{ChamferDistance}(\mathcal{P},\mathcal{P}_\text{project}).
\end{equation}

Meanwhile, we devise a direct constraint on UDF for an intra-consistency by minimizing the mean UDF value of all points of the input point cloud:
\begin{equation}
    L_{\text{intra}}=\sum_{p \in \mathcal{P}} \text{UDF}(p),
\end{equation}
where $\text{UDF}(\cdot)$ is the UDF value, it is computed by taking the L2 norm of the estimated UDF projection flow.

Imposing the above constraints would lead to noisy UDFs where the monotonicity of UDF at one side of the zero-value surface may be broken. To this end, we introduce an extra loss to ensure the monotonicity of the UDF (Fig.~\ref{fig:method_illustration} b).
For each upsampled point $q \in \mathcal{P}_\text{upsample}$, we compute its offset point $q^{\prime}$ by moving $q$ along UDF gradient for a distance of half of $q$'s UDF value. $q^{\prime}$ is expected to reside at the same side as $q$ about the surface. The UDF gradient at the two points should be the same and the UDF value at $q^{\prime}$ should be half of that of $q$.
We feed $q$ and $q^{\prime}$ into the UDF projection flow estimation network, respectively. The network is trained in a siamese fashion.
It optimizes the inter-consistency between the two estimated UDFs:
\begin{eqnarray}
\begin{aligned}
   L_{\text{inter}}=\sum_{q \in \mathcal{P}_\text{upsample}} &(\left|0.5\cdot\text{UDF}(q)-\text{UDF}(q^{\prime})\right|\\
   +&\Vert\Delta\text{UDF}(q)-\Delta\text{UDF}(q^{\prime})\Vert _2),
\end{aligned}
\label{eq:udf_grad}
\end{eqnarray}
where $\Delta\text{UDF}(\cdot)$ is the UDF projection flow.
To sum up, the total loss is: $L = w_{\text{chamfer}}L_{\text{chamfer}} +
w_{\text{inter}}L_{\text{inter}} + w_{\text{intra}}L_{\text{intra}}$,
where $w_{\text{chamfer}}$, $w_{\text{intra}}$, and $w_{\text{inter}}$ are the pre-defined weights. We set $w_{\text{chamfer}}=10$, $w_{\text{intra}}=1$, and $w_{\text{intra}}=1$,


\subsection{Learning-based Mesh Extraction}\label{sec:iso_ext}
Next, we describe how to extract the mesh surface based on the estimated UDF. The marching cube algorithm has proven to be extremely effective in extracting iso-surface from signed distance fields (SDFs).
However, since no sign is provided to distinguish the inner and outer of the surface by UDFs, directly applying the conventional marching cube algorithm is infeasible.
Therefore, we propose a learning-based UDF sign estimation method to solve this problem. CAP~\cite{capudf} also provide a method to reconstruct mesh, they have the sign conflict problem in theory.

The method starts by dividing the 3D space into $H^3$ voxels and calculating the UDF value and UDF gradient direction of every vertex.
For each $h^3$ ($h=5$) regular grid of overlapping local regions, we propose a local sign estimation network to predict the sign of the vertices.
To be specific, as Fig. \ref{fig:tiny_net} shown, the network backbone is stacked by several 3D Convolution layers for UDF-sensitive feature extraction. By feeding the estimated UDF and positional encoding~\cite{2020Fourier} of the grid vertices into the network,
we make binary classifications for each vertex, indicating the sign of UDFs.
Note that only the relative sign of the UDFs is needed. As a result, we train the network using a modified binary cross-entropy loss:
\begin{align}
    L = \sum_{c \in \mathcal{C}}min(\sum_{v \in \mathcal{V}}\text{BC}(y^{c}_v, \Tilde{y}^{c}_v), \sum_{v \in \mathcal{V}}\text{BC}(-y^{c}_v,\Tilde{y}^{c}_v)),
\end{align}
where $\mathcal{C}$ is the grid set, $\mathcal{V}$ is vertex set. $\text{BC}$ is the binary cross-entropy loss function, $y^{c}_v$ is the predicted sign at vertex $v$ of grid $c$, and $\Tilde{y} \in \{0, 1\}$ is the ground-truth sign.
In practice, we fix the sign of one corner and let the network predict the rest.

Note that, since the network is trained for local sign prediction, it is easy to train and generalizes well across different shapes.
We show the cross-category and cross-dataset generality of our method in the ablation study. 
\subsection{Discussion}

\paragraph{Approaching Vector Explanation}
First, we need to explain why we need approaching vector and why we can learn the approaching vector. The mechanism of approaching vector is shown in Fig. \ref{fig:shift_error} (a). Let's imagine a local plane $P$ with ground-truth  normal $n$. On the one hand, if the approaching vector is on the line of the ground-truth normal, we can project the points around the surface, according to Eq. (\ref{eq:shift1}), to the plane $P$. On the other hand, if the approaching vector is not on the line of the ground-truth normal, the points around the surface will not be projected to the surface and the chamfer loss will be large. Thus, if we project the points around the surface according to Eq. (\ref{eq:shift1}), the chamfer loss will push the approaching vector close to the point normal. According to Eq. (\ref{eq:shift1}), if we flip the orientation of the approaching vector, the result shift $s_1$ do not change. Because $s_1$ is the even function of the approaching vector. Thus, it is possible that our approaching vector can be close to the normal. We also provide experiments validation in Sec. \ref{sec:stage2}.

\subsection{Difference between CAP and Neural Pull}
In the field (SDF or UDF) prediction stage, the two previous methods do not need the label of query point. It is Neural Pull~\cite{2020npull} and CAP~\cite{capudf}. The former is the pioneer work designed for SDF prediction. The latter is designed for UDF prediction. Our method is based on UDF and do not need label of query point as well. Therefore, we compare the difference between ours and the two previous works.

First, our method and CAP~\cite{capudf} can represent open surface and close surface. While Neural Pull~\cite{2020npull} can only represent close surface. The reason is that ours and CAP is based on UDF while Neural Pull is based on SDF. Second, Neural Pull and CAP have to train the network for every shape which is time-consuming. While ours can predict the UDF fast at inference stage which is much faster. The reason is that we learn a prior with input as condition, similar to Occupancy Network~\cite{2019Occupancy}. While CAP and Neural Pull adopt a pipeline similar to SAL~\cite{SAL}. The pipeline of CAP~\cite{capudf}, Neural Pull~\cite{SAL} and SAL~\cite{SAL} is easier to obtain continuous SDF or UDF, because neural network is prone to be a continuous function with the cost of longer time at inference stage . Third, due to the existence of regularization Eq. (\ref{eq:udf_grad}), we can make the UDF near surface more reasonable, thus ours can work better on relatively sparse point cloud. While Neural Pull and CAP need dense point as input. This is shown in Tab. \ref{tab:diff}. The total time consumption can be found in Appendix.

\begin{table}[]
\centering
\begin{tabular}{llll}
\toprule
Method         & Open surface & Inference Time & Sparse Input\\
\midrule
NP~\cite{2020npull}  & No    & 29min22s  & Hard   \\
CAP~\cite{capudf} & Yes     & 21min56s  & Hard   \\
Ours & Yes     & 9s &Easy     \\

\bottomrule
\end{tabular}
\vspace{3pt}
\caption{The UDF estimation comparison between NP~\cite{2020npull}, CAP~\cite{capudf} and ours.}
\label{tab:diff}
\end{table}

\subsection{Details of of UDF Estimation Network}
\paragraph{Network} The network is composed of 4 Point Transformer layers~\cite{zhao2021pointtrans}, the output width of each layer is $32$, $128$, $256$ and $256$. The number of k-nearest points is $36$. Through the 4 layers, we can obtain per-point feature. Then we can use a MLP to predict the approaching vector and calculate the shifted points with $s_1$ according to Eq. (\ref{eq:shift1}). Next, we calculate the refined shift $s_2$ with a learning based interpolation method in Eq. (\ref{eq:interpolate}) followed by MLP.

\paragraph{Implementation Details}  
During training, we adopt cosine learning rate adjustment~\cite{2016coslr}, the initial learning rate is 0.001. We use Adam~\cite{adam} optimizer to learn the parameter. The batch-size is 2. The query point for training is obtained by randomly sampling point around the original point with the uniform distribution $U(-0.03, 0.03)$. During testing, to obtain the UDF in the whole space, we first partition the space into $256^3$ voxels. We use the network to predict the UDF of the voxel center.

%% file: figures/network.tex
\begin{figure} 
\centering
  \includegraphics[width=1\columnwidth]{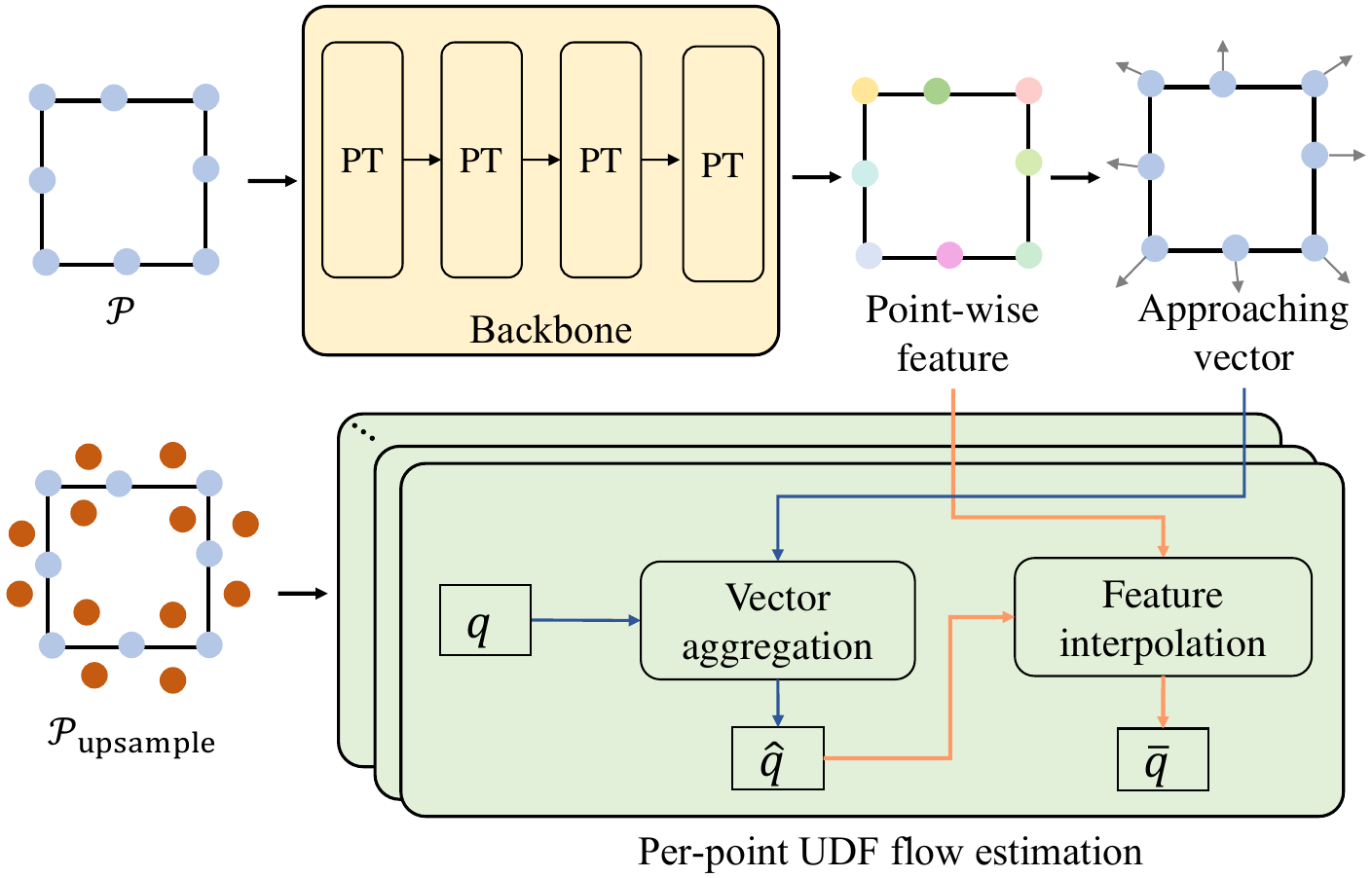}
  \caption{The UDF estimation network. Given the input point cloud $\mathcal{P}$ and the upsampled point cloud $\mathcal{P}_\text{upsample}$, the network estimates the per-point UDF projection flow in two stages with the guidance of the predicted approaching vector and the point-wise feature, respectively.}
  \label{fig:network}
\end{figure}


%% file: figures/method_illustration.tex
\begin{figure}
\centering
  \includegraphics[width=1.0\columnwidth]{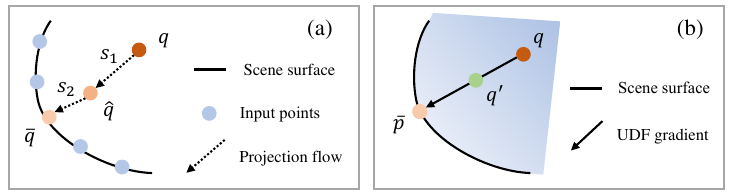}
  \caption{(a): The two-stage UDF projection flow prediction. The projection vectors $s_1$ and $s_2$ are predicted respectively by aggregating features at different scales. (b): The inter-consistency loss that ensures the monotonicity of the UDF.}
  \label{fig:method_illustration}
\end{figure}

%% file: figures/tiny_net.tex
\begin{figure} 
\centering
  \includegraphics[width=\columnwidth]{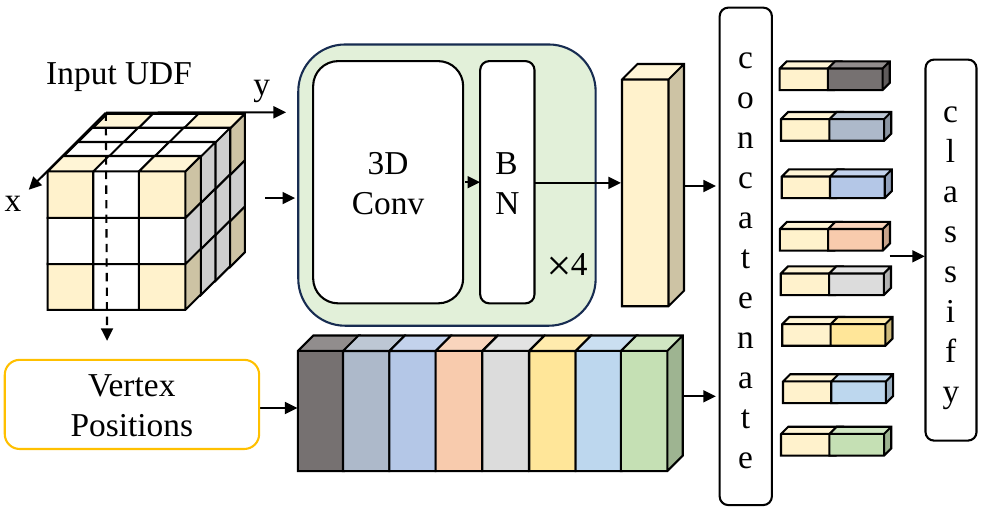}
  \caption{The architecture of the local UDF sign estimation network.}
  \label{fig:tiny_net}
\end{figure}

%% file: 4exp.tex
\section{Experiments}
\subsection{Experiments Settings}

\paragraph{Benchmarks} We evaluate our method on 3 widely used open-source datasets. The first is ShapeNet~\cite{chang2015shapenet}, we random choose 1300 shapes from 13 commonly used categories, every categories with 100 shapes. The second is the MGN dataset, including pants and shirts. The third ScanNet~\cite{dai2017scannet}, a real scene from scanning, we choose 100 scenes from it. To comprehensively compare our results with other methods, we make use of the most commonly used metrics in surface reconstruction for comparison. $CD_1$ is Chamfer-$L_1$, a set operation which measures distance of two point sets.
In short, we show how we make use of chamfer distance to constrain the training. $CD_1$ equation is shown as the Eq.(\ref{eq:chamfer}),
\begin{equation}
\label{eq:chamfer}
\begin{aligned}
\mathrm{CD}_{1}=& \frac{1}{2 N_{x}} \sum_{i=1}^{N_{x}}\left\|\mathbf{x}_{i}-\mathcal{S}_{y }\left(\mathbf{x}_{i}\right)\right\|_1+\\
& \frac{1}{2 N_{y}} \sum_{i=1}^{N_{y}}\left\|\mathbf{y}_{i}-\mathcal{S}_{x}\left(\mathbf{y}_{i}\right)\right\|_1,
\end{aligned}
\end{equation}
$\{ \mathbf{x}_{i}, i=0\cdots N_x\}$ is the points set sampled from ground-truth mesh. $\{\mathbf{y}_{i}, i=0\cdots N_y\}$ is the point set sampled from reconstructed mesh. $\mathcal{S}_{y}(\mathbf{x}_{i})$ means the nearest point to $\mathbf{x}_{i}$ in reconstructed point set. $\mathcal{S}_{x}(\mathbf{y}_{i})$ means the nearest point to $\mathbf{y}_{i}$ in ground-truth point set. $\|\cdot\|_1$ means $L1-distance$.


\subsection{Close Surface Reconstruction}
This part is a standard implicit surface reconstruction task, lots of previous works~\cite{2019DeepSDF, 2020Convolutional, 2019Occupancy, 2020ndf} have adopted the same setting. 
We first evaluate our method by close shape reconstruction on ShapeNet~\cite{chang2015shapenet}.
In detail, we uniformly sample 3000 points on the watertight mesh as input. Our method is self-supervised method, as well as Neural Pulling~\cite{2020npull} and CAP~\cite{capudf} in Fig. \ref{fig:shapenet_result} and Fig. \ref{fig:shapenet_result2}. Others are supervised methods, which need ground-truth label of query point. As for PSR~\cite{2013Poisson}, we provide ground-truth normal as input. We can see that our method achieves state-of-the-art performance in almost all the 13 classes. SuperUDF can also outperform the supervised methods since our method can extract more implicit geometry information on less supervision, such as the UDF projection flow, for the surface reconstruction learning. 
\begin{table}[]
    \centering
\begin{tabular}{llllll}
\toprule
Class     & GIFS            & CAP    & NPull & ConvOcc & Ours            \\
& ~\cite{gifs}  &~\cite{capudf} &~\cite{2020npull} &~\cite{2020Convolutional}&\\
\midrule
airplane  & 0.0026          & 0.0031 & 0.0037 & 0.0043  & \textbf{0.0022} \\
bench     & 0.0058          & 0.0059 & 0.0079 & 0.0048  & \textbf{0.0029} \\
cabinet   & 0.0104          & 0.0048 & 0.0067 & 0.0065  & \textbf{0.0039} \\
car       & 0.0052          & 0.0054 & 0.0060 & 0.0081  & \textbf{0.0029} \\
chair     & 0.0041          & 0.0042 & 0.0122 & 0.0089  & \textbf{0.0039} \\
display   & 0.0055          & 0.0047 & 0.0049 & 0.0052  & \textbf{0.0034} \\
lamp      & 0.0078          & 0.0082 & 0.0089 & 0.0153  & \textbf{0.0030} \\
speaker   & 0.0081          & 0.0063 & 0.0075 & 0.0080  & \textbf{0.0041} \\
rifle     & 0.0029 & 0.0013 & 0.0016 & 0.0031  & \textbf{0.0011} \\
sofa      & 0.0069 & 0.0050 & 0.0051 & 0.0053  & \textbf{0.0033} \\
table     & 0.0065          & 0.0088 & 0.0117 & 0.0050  & \textbf{0.0025} \\
phone & 0.0049          & 0.0024 & 0.0028 & 0.0034  & \textbf{0.0021} \\
vessel    & 0.0033          & 0.0025 & 0.0037 & 0.0050  & \textbf{0.0021} \\
\midrule
Mean      & 0.0057          & 0.0048 & 0.0063 & 0.0064  & \textbf{0.0028}\\
\bottomrule
\end{tabular}
\vspace{3pt}
    \caption{$CD_1$ comparison on ShapeNet within 13 classes}
    \label{tab:my_label}
\end{table}

\input{figures/shapenet_result.tex}

\subsection{Open Surface Reconstruction}
Since our pipeline adopts UDF to represent the implicit surface, SuperUDF can reconstruct open surface as well. We uniformly sample 3000 points on each MGN mesh, and reconstructing the implicit surface. SDF-based methods cannot represent the garment correctly because it is open surface. The quantitative result is shown in Tab. \ref{tab:garment}, we can see that our method outperforms the supervised method NDF~\cite{2020ndf}, PSR~\cite{2013Poisson}, CAP~\cite{capudf} and GIFS~\cite{gifs}. The visualization result is shown in Fig. \ref{fig:garment_result}. We can see that our reconstructed mesh is closer to the ground-truth and retain more fine-grained details. 

\begin{table}[]
\centering
\begin{tabular}{lllll}
\toprule
\#points      & CAP~\cite{capudf}      & NDF~\cite{2020ndf}      & GIFS~\cite{gifs}      & Ours     \\
\midrule
1000 & 0.0047 & 0.0062 & 0.0041 & \textbf{0.0036} \\
3000 & 0.0035 & 0.0038 & 0.0033 & \textbf{0.0024} \\
\bottomrule
\end{tabular}
\vspace{3pt}
\caption{$CD_1$ comparison results on MGN dataset.}
\label{tab:garment}
\end{table}

\begin{table}[]
\centering
\begin{tabular}{lllll}
\toprule
\#points       & CAP~\cite{capudf}      & NDF~\cite{2020ndf}      & GIFS~\cite{gifs}      & Ours     \\
\midrule
10000  & 0.0044 & 0.0047 & 0.0043 & \textbf{0.0039} \\
3000  & 0.0075 & 0.0096 & 0.0068 & \textbf{0.0057}\\
\bottomrule
\end{tabular}
\vspace{3pt}
\caption{$CD_1$ comparison on ScanNet with different number of sample point.}
\label{tab:scan}
\end{table}

\input{figures/garment_result.tex}

\subsection{Real Scene Reconstruction}
Real scene reconstruction is more challenging since the input point cloud is usually open and incomplete, while there is no reconstruction supervision for training. Our self-supervised method can train the network with the raw point cloud directly and does not rely on the synthetic dataset for supervision.
We evaluate our method on ScanNet~\cite{dai2017scannet}. For each scene, we sample 3000 or 10000 points as input. The visualization result is shown in Fig. \ref{fig:scan_result}. By comparing with the reconstructed ground truth, we can see that our method can achieve good result with fine-grained details. As for quantitative result, we sample 100000 points on our mesh and ground truth mesh, then calculate the $CD_1$ as the metric. The quantative result is shown in Tab. \ref{tab:scan}.

We provide some reasons of artefacts and an improvement direction. The artefacts at least comes from two aspects, the local discontinuities of the UDF gradient near the surface and the noise near the edge of open surface. For the first problem, it is due to the sudden UDF gradient direction changing near the surface, which is tightly connected with the UDF representation way. Therefore, the artefacts might be more than SDF-based method. However, UDF representation is still a very promising direction and the artefacts can be removed with the UDF accuracy increase. As for the second problem, the UDF gradient from two sides of the surface is not completely opposite to each other on the edge of the surface, it will lead to some noise during mesh extraction. To solve the second problem, we can design a half plane rather than a complete plane in Eq(\ref{eq:shift1}) to fit the edge of the open surface.

\input{figures/scan_result.tex}

\input{figures/ab_result.tex}

\input{figures/shapenet_result2.tex}

\subsection{Validation Experiments}
\label{sec:stage2}
\paragraph{The Explanation of 2 stage Learning Method}
\input{figures/shift_error.tex}
In fact, only the stage 1 and stage 2 can predict the UDF flow alone. Here, we analyze why we need the 2-stage pipeline.
Let's consider if the local plane is a curve surface in Fig. \ref{fig:shift_error} (b). Then the shift $s_1$ according to Eq. (\ref{eq:shift1}) will have a systematical error. The points around the surface will not be projected to the surface. Thus, we need a refinement stage to push the points near the surface to the surface exactly. Next, we explain why we do not use the second stage multiple times. In theory, neural network can fit any curve thus can move the query point to any curve. However, according to our experiments, the neural network is prone to move the query point to the input point cloud rather than evenly scattering the query point to the surface, as shown in second column in Fig. \ref{fig:stage}. If the query point moves to the position of input point cloud only, the network is over-fitting and the predicted UDF is obviously wrong. In summary, we cannot use one stage multiple times. In the meanwhile, two-stage method can help to reduce the systematic error in Fig. \ref{eq:shift1} (b) and avoid the query point moving to the position of input point cloud only. It is shown in Fig. \ref{fig:stage}. We can see that better upsampling result can be reached with 2 stages together.

\input{figures/stage}

We provide the ablation study results in Tab. \ref{tab:stage}. We can see that stage-1 + stage-2 can outperform both stage-1 and stage-1+stage-2. In experiments, it proves the necessity of the two stage method. 

\begin{table}[!t]
  \centering
  \scalebox{0.85}{
  \begin{tabular}{l|l|l|l}
    \toprule
     & Only Stage-1 & Stage-1 repeat 2 times & Stage-1 + stage-2\\
    \midrule
    airplane & 0.0025 & 0.0024 & \bf{0.0022}\\
    bench & 0.0032 & 0.0031 & \bf{0.0029}\\
    cabinet& 0.0044 & 0.0044 & \bf{0.0039}\\
    car & 0.0036 & 0.0033 & \bf{0.0029}\\
    chair & 0.0043 & 0.0041 & \bf{0.0039}\\
    display & 0.0037 & 0.0036 & \bf{0.0034}\\
    lamp & 0.0034 & 0.0033 & \bf{0.0030}\\
    speaker & 0.0043 & 0.0043 & \bf{0.0041}\\
    rifle& 0.0012 & 0.0011 & \bf{0.0011}\\
    sofa & 0.0036 & 0.0034 & \bf{0.0033}\\
    table & 0.0027 & 0.0026 & \bf{0.0025}\\
    phone & 0.0023 & 0.0023 & \bf{0.0021}\\
    vessel& 0.0025 & 0.0024 & \bf{0.0021}\\
    Mean & 0.0032 & 0.0031 & \bf{0.0028}\\
    MGN & 0.0027 & 0.0026 & \bf{0.0024}\\
    ScanNet & 0.0044 & 0.0041 & \bf{0.0039}\\
    \bottomrule
  \end{tabular}
  }
  \vspace{3pt}
  \caption{Ablation study of second-stage prediction.}
      \label{tab:stage}
\end{table}

\paragraph{Approaching Vector Explanation} What's more, we provide the comparison result of approaching vector and ground truth normal visualization in Fig. \ref{fig:normal}. we can see that the visualization result is close to the ground-truth normal. What's more, we provide the quantitative results as well. The result is shown in Tab. \ref{tab:nc}. we can see that our approaching vector is close to the ground-truth normal. 
\input{figures/normal.tex}

\begin{table}[]
\centering
\begin{tabular}{llll}
\toprule
Dataset & ShapeNet & MGN & ScanNet \\
\midrule
NC      & 0.9023   & 0.8875  & 0.8934 \\
\bottomrule
\end{tabular}
\vspace{3pt}
\caption{The normal consistency between approaching vector and ground-truth normal.}
\label{tab:nc}
\end{table}


\subsection{Robust Analysis of $k$ in Stage-1}
In the UDF projection flow estimation part, we propose a 2-stage way to predict the query point shift. In Eq. (\ref{eq:shift1}), $k$ is an important parameter. Here, we provide the robust analysis of $k$ in Tab. \ref{tab:k_ablation}. We can see that when $k$ varies, the result is robust.

\begin{table}[]
\centering
\begin{tabular}{l|lll}
\toprule
Loss & ShapeNet & MGN  & ScanNet \\
\midrule
k=8    & 0.0028       & 0.0024 & 0.0039   \\
k=16   & 0.0029       & 0.0026 & 0.0041   \\
k=32    & 0.0028       & 0.0031 & 0.0040   \\
k=64    & 0.0030       & 0.0025 & 0.0041   \\
\bottomrule
\end{tabular}
\vspace{3pt}
\caption{The $CD_1$ on 3 datasets when $k$ in Eq. (\ref{eq:shift1}) varies.}
\label{tab:k_ablation}
\end{table}

\subsection{Ablation Study of the Regularization Term}
We verify the effectiveness of the proposed inter-consistency loss in Fig. \ref{fig:ab_result}. We show the visualization result of up-sampled point cloud w/o the regularization term and 2D slice of UDF. By comparing to result w/o the regularization, we can see that our result is obviously better with the regularization.

Due to the inter-loss which can regularize the UDF distribution around the surface, our self-supervised method can work on rather sparse input, otherwise the query point will shrink to the input point cloud, as shown in Fig. \ref{fig:ab_result}, thus the UDF will have lots of errors.

Here we provide the result w/ and w/o inter-loss with different number of point cloud as input. By comparison, we can see that method with regularization term perform much better than method without regularization term. Meanwhile, we also compare ours with other methods with different number of point cloud as input, shown in Tab. \ref{tab:sparse}. We can see that our method with regularization can work well on both sparse and dense point cloud. The two comparisons show the effectiveness of the regularization term.

As for the reason why our method without regularization term is worse than the CAP~\cite{capudf} and Neural Pull~\cite{2020npull}, because the predicted UDF in our pipeline is based on local feature, while NP and CAP model one shape with one neural network, thus ours are harder to obtain continuous UDF or SDF without regularization. However the cost of their pipeline is more time at inference stage. 

    

\subsection{Mesh Extraction Comparison}
In order to compare the performance of our mesh extraction part and the MeshUDF~\cite{2021MeshUDF}, we design two experiments, mesh extraction on ground truth UDF (clean) and mesh extraction on predicted UDF (noisy). The UDF is predicted by our UDF prediction network. In order to show that the good performance is not only on one dataset, we show the result on 2 datasets, ShapeNet, MGN. Specifically in Tab. \ref{tab:compare_meshudf}, we can see that our method show better performance than MeshUDF on both datasets and on both GT and Predicted UDF. What’s more, we provide the visualization result of the mesh extracted my ours and MeshUDF in Fig. \ref{fig:compare_meshudf}. It is consistent with the result on Tab. \ref{tab:compare_meshudf}. All experiments is with resolution $256^3$.
\begin{table}[]
    \centering
    \begin{tabular}{l|ll|ll}
\toprule & \multicolumn{2}{c}{\text { GT UDF }} & \multicolumn{2}{|c}{\text { Predicted UDF }} \\
\midrule \text { Dataset } & \text { ShapeNet(V) } & \text { MGN } & \text { ShapeNet(V) } & \text { MGN } \\
 \midrule \text { MeshUDF } & 0.0033 & 0.0024 & 0.0035 & 0.0027 \\
 \text { Ours } & \textbf{0.0018} & \textbf{0.0022} & \textbf{0.0021} & \textbf{0.0024} \\
\bottomrule
\end{tabular}
\vspace{3pt}
    \caption{The comparison between our method and MeshUDF~\cite{2021MeshUDF}. ShapeNet(V) means vessel class of ShapeNet.}
    \label{tab:compare_meshudf}
\end{table}
\input{figures/compare_meshudf}
 \subsection{Generalization of Iso-Surface Extraction}
We define the metric acc for evaluating the generalization ability. Detail definition is in Appendix. We only choose the cubes near the surface for mesh extraction. In detail, the distance from the center of the cube to the nearest point in the point cloud should be less than $0.03$.

We demonstrate that our method can be trained and tested on different datasets as shown in Tab. \ref{tab:crossdataset}. We can see that the test accuracy will not decrease much if the network is trained on a different small-scale dataset. The reason is the repeatability of micro structures as discussed.

\begin{table}[]
\centering
\begin{tabular}{llll}
\toprule
train dataset & test dataset & \#train shape & accuracy \\
\midrule
ShapeNet      & ScanNet      & 5                       & 0.95     \\
ShapeNet      & ScanNet      & 20                       & 0.96     \\
ScanNet       & ShapeNet     & 5                       & 0.97     \\
ScanNet       & ShapeNet     & 20                       & 0.98    \\
\bottomrule

\end{tabular}
\vspace{3pt}
\caption{The accuracy training on different number of shapes and test on different dataset.}
\label{tab:crossdataset}
\end{table}

\subsection{More Result of Surface Reconstruction}

We also provide the result on FAMOUS dataset released by Points2Surf~\cite{2020Points2Surf}. The result is shown in Fig. \ref{fig:famous_result}. We can see that our method can reconstruct the surface with more geometry details.

\input{figures/famous_result.tex}

\begin{table}[]
\begin{tabular}{l|l|l|lll}
\toprule
\multirow{5}{*}{ShapeNet} & \#points             & 1000            & 3000            & 10000           & 20000           \\
\midrule
                          & NPull               & 0.0075          & 0.0060          & 0.0047          & 0.0027          \\
                          & CAP                 & 0.0072          & 0.0048          & 0.0026          & \textbf{0.0023} \\
                          & Ours w/o inter-loss & 0.0122          & 0.0082          & 0.0052          & 0.0028          \\
                          & Ours w inter-loss   & \textbf{0.0039} & \textbf{0.0028} & \textbf{0.0024} & \textbf{0.0023} \\
                          \midrule
\multirow{3}{*}{MGN}      & CAP                 & 0.0047          & 0.0035          & 0.0022          & \textbf{0.0018} \\
                          & Ours w/o inter-loss & 0.0076          & 0.0052          & 0.0026          & 0.0023          \\
                          & Ours w inter-loss   & \textbf{0.0036} & \textbf{0.0024} & \textbf{0.0021} & 0.0019          \\
                          \midrule
\multirow{3}{*}{ScanNet}  & CAP                 & -               & 0.0075          & 0.0044          & \textbf{0.0035} \\
                          & Ours w/o inter-loss & -               & 0.0097          & 0.0074          & 0.0056          \\
                          & Ours w inter-loss   & -               & \textbf{0.0057} & \textbf{0.0039} & 0.0036 \\
                          \bottomrule
                          
\end{tabular}
\vspace{3pt}
\caption{$CD1$ comparison results with different sparse level on ShapeNet, MGN and ScanNet.}
\label{tab:sparse}
\end{table}

    

    
    


%% file: figures/shapenet_result.tex
\begin{figure*} 
\centering
  \includegraphics[width=2\columnwidth]{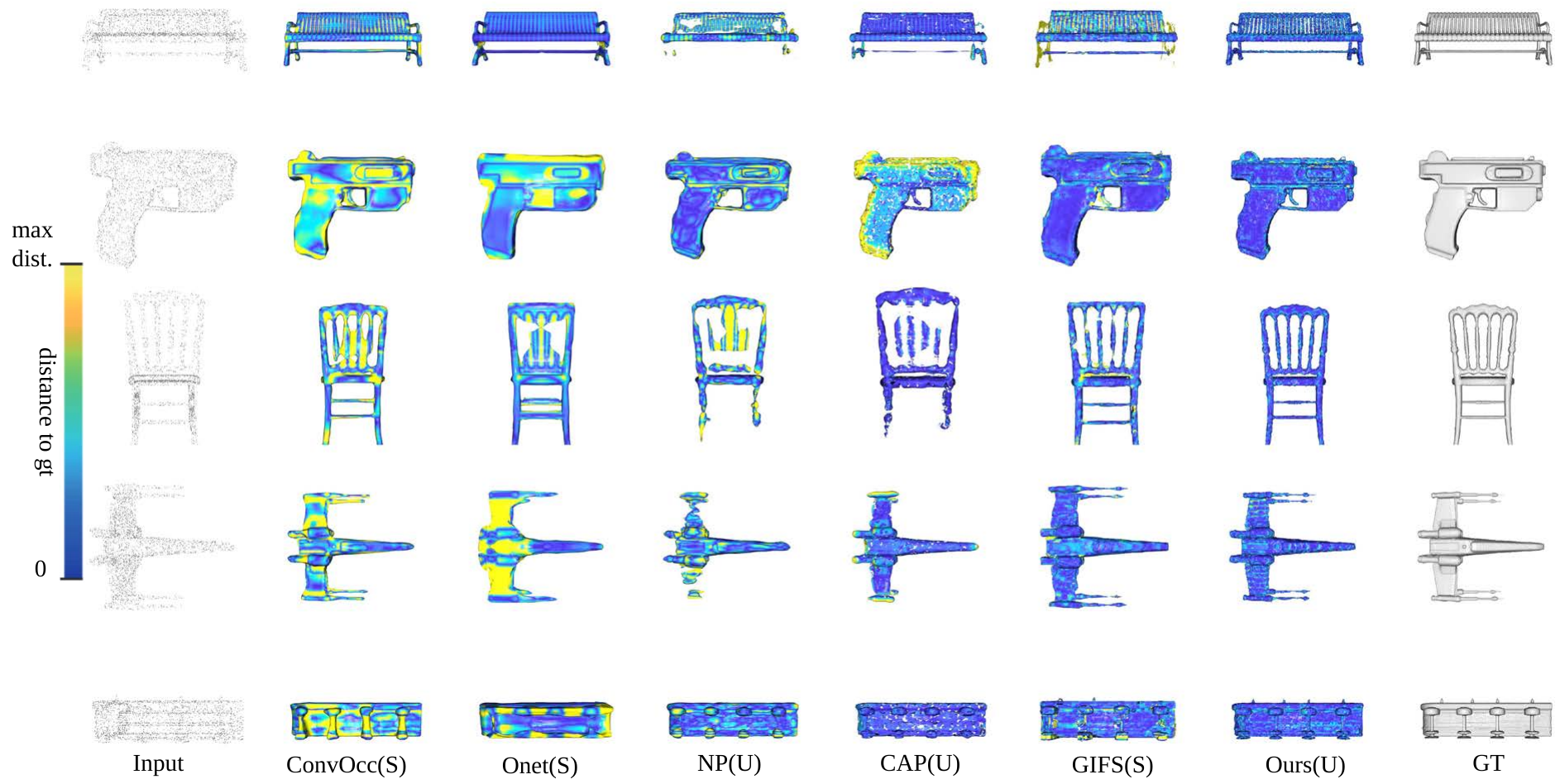}
  \caption{Visualized results of our method and state-of-the-art alternatives. Note that we provide the ground-truth normal as input for PSR as it requires. The color map in the figure is the error distribution of the reconstruction. Blue indicates low reconstruction error while yellow indicates higher error values. (S) mean supervised method and (U) means unsupervised method. Methods are ConvOcc\cite{2020Convolutional}, Onet\cite{2019Occupancy}, NP\cite{2020npull}, CAP\cite{capudf}, GIFS\cite{gifs}. Max dist is 0.017. The number of input point cloud is 3000.}
  \label{fig:shapenet_result}
\end{figure*}

%% file: figures/garment_result.tex
\begin{figure} 
\centering
  \includegraphics[width=\columnwidth]{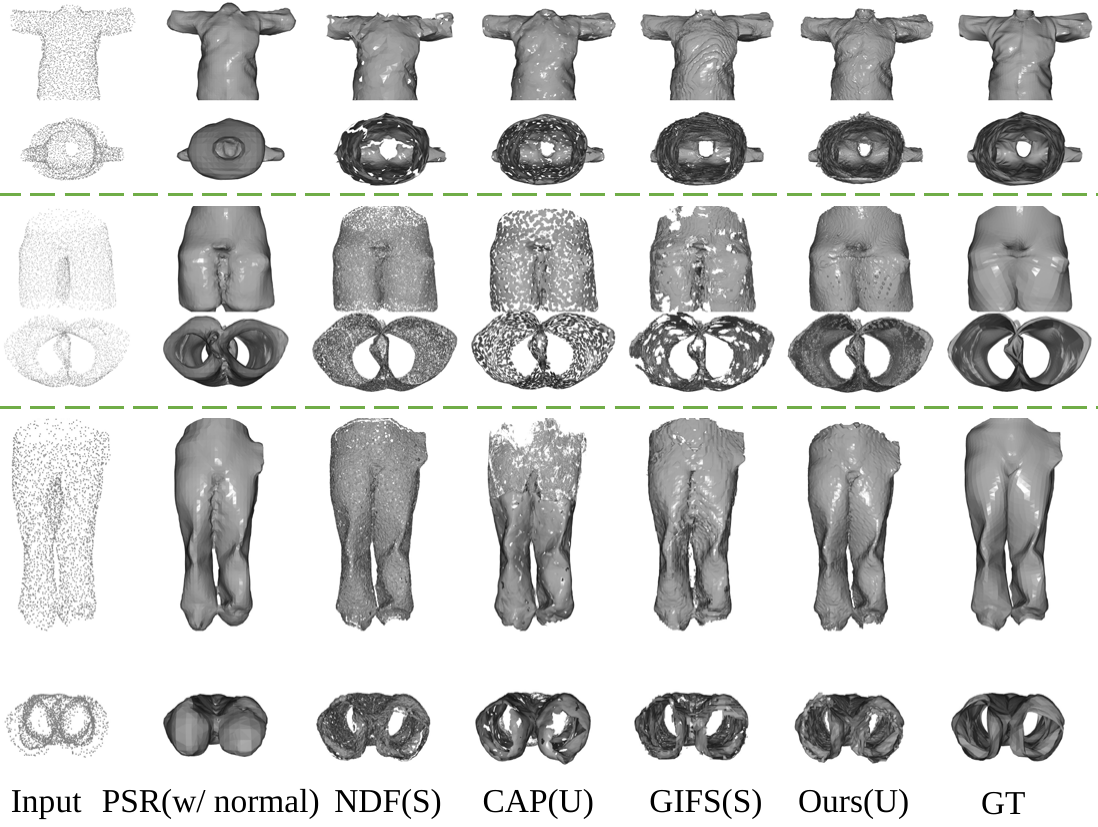}
  \caption{Visualized results of our method, traditional reconstruction methods PSR\cite{2013Poisson} and 3 deep method NDF\cite{2020ndf}, CAP\cite{capudf}, GIFS\cite{gifs} on garment dataset. Note that we provide the ground-truth normal as input for PSR as it requires. The number of input point cloud is 3000.}
  \label{fig:garment_result}
\end{figure}

%% file: figures/scan_result.tex
\begin{figure*} 
\centering
  \includegraphics[width=2\columnwidth]{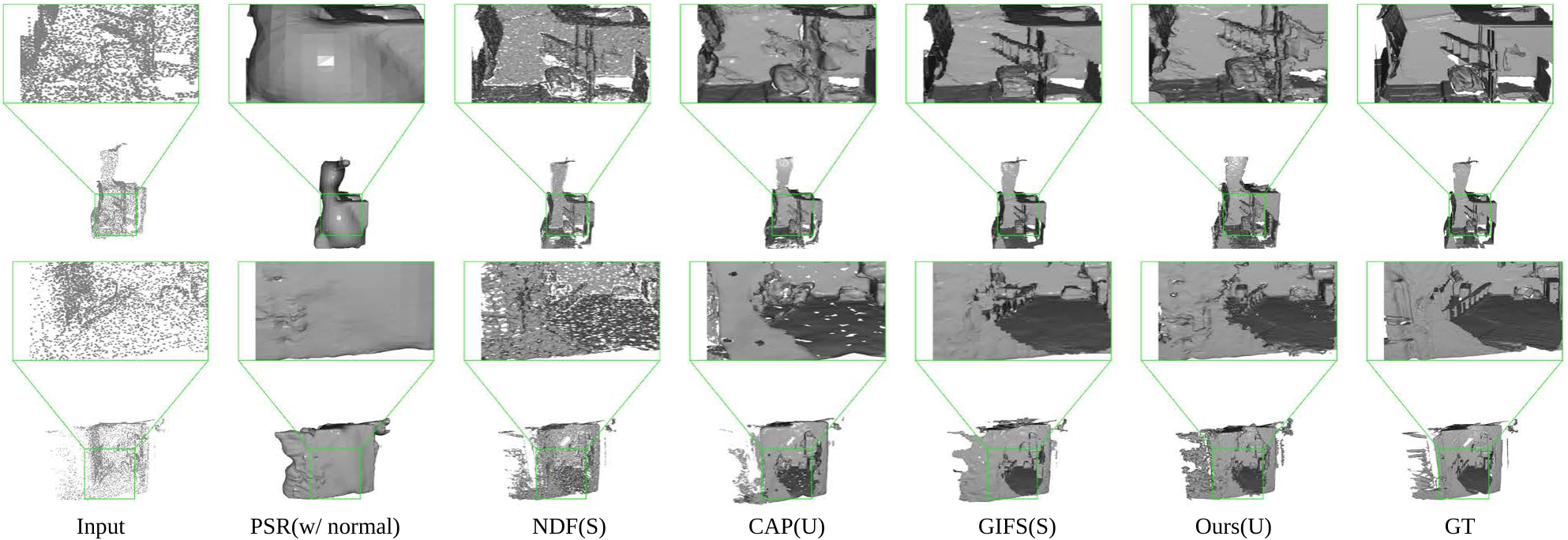}
  \caption{Visualized results of our method and NDF\cite{2020ndf}, CAP\cite{capudf} and GIFS\cite{gifs}. The number of input point cloud is 10000.}
  \label{fig:scan_result}
\end{figure*}

%% file: figures/ab_result.tex
\begin{figure} 
\centering
  \includegraphics[width=0.8\columnwidth]{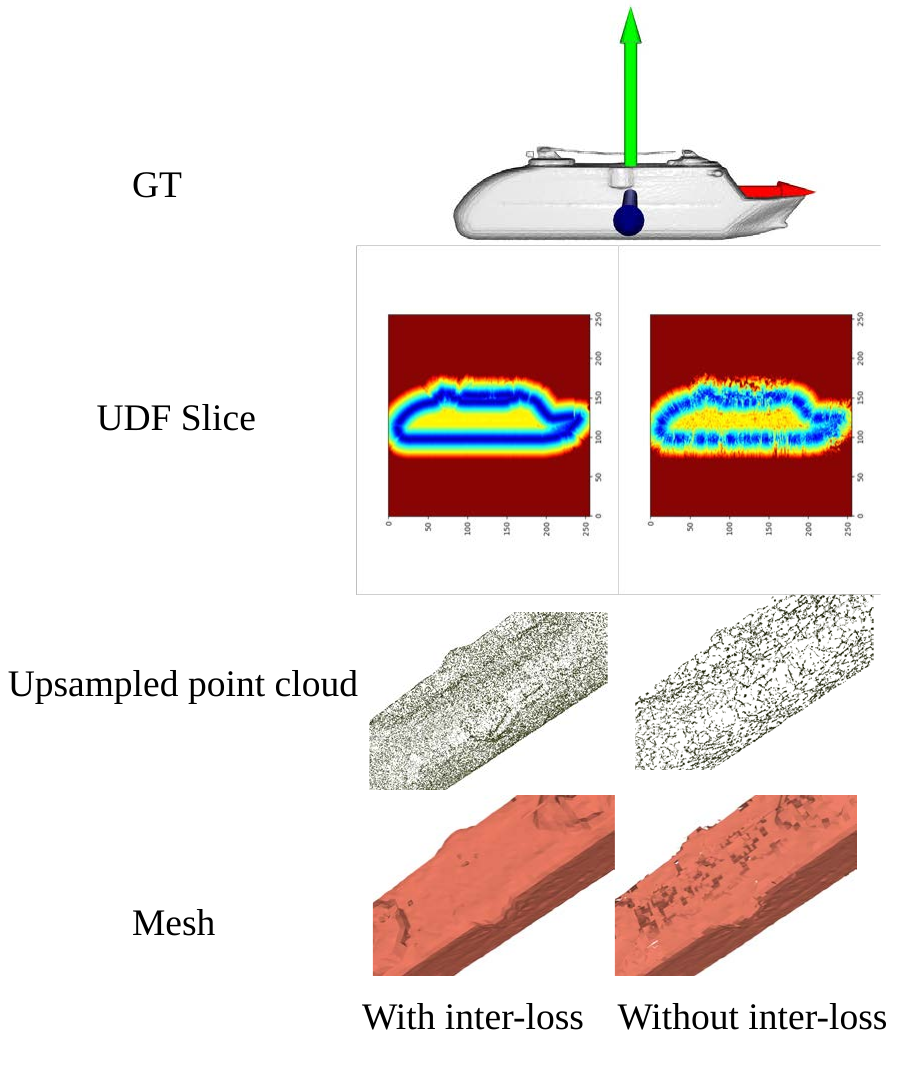}
  \caption{Visualized results of the ablation study on the regularization term. The proposed inter-consistency loss can significantly improve the up-sample and reconstruction quality. }
  \vspace{-10pt}
  \label{fig:ab_result}
\end{figure}

%% file: figures/shapenet_result2.tex
\begin{figure*} 
\centering
  \includegraphics[width=2\columnwidth]{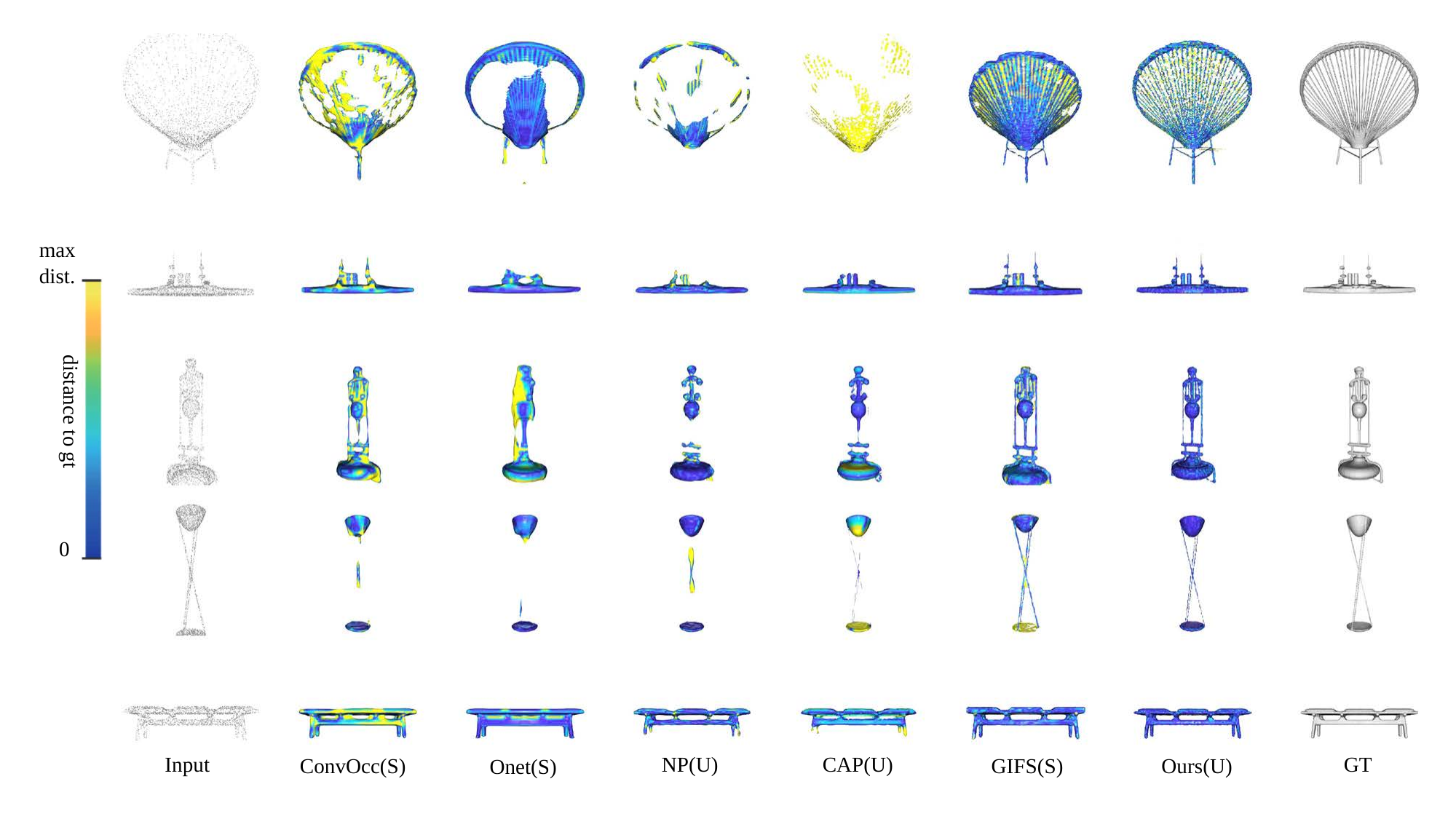}
  \caption{Visualized results of our method and state-of-the-art alternatives. Note that we provide the ground-truth normal as input for PSR as it requires. The color map in the figure is the error distribution of the reconstruction. Blue indicates low reconstruction error while yellow indicates higher error values. (S) mean supervised method and (U) means unsupervised method. Methods are ConvOcc\cite{2020Convolutional}, Onet\cite{2019Occupancy}, NP\cite{2020npull}, CAP\cite{capudf}, GIFS\cite{gifs}.}
  \label{fig:shapenet_result2}
\end{figure*}

%% file: figures/shift_error.tex
\begin{figure}
\centering
  \includegraphics[width=\columnwidth]{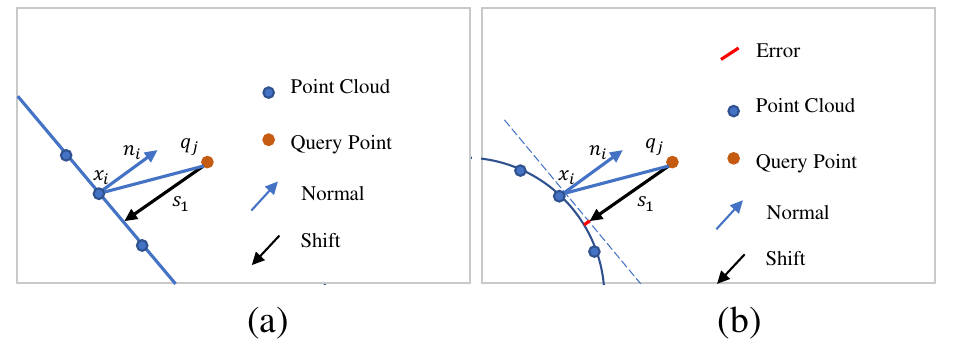}
  \caption{The mechanism of $s_1$ is in (a). The systematic error of $s_1$ is in (b).}
  \label{fig:shift_error}
\end{figure}

%% file: figures/stage.tex
\begin{figure}
\centering
  \includegraphics[width=\columnwidth]{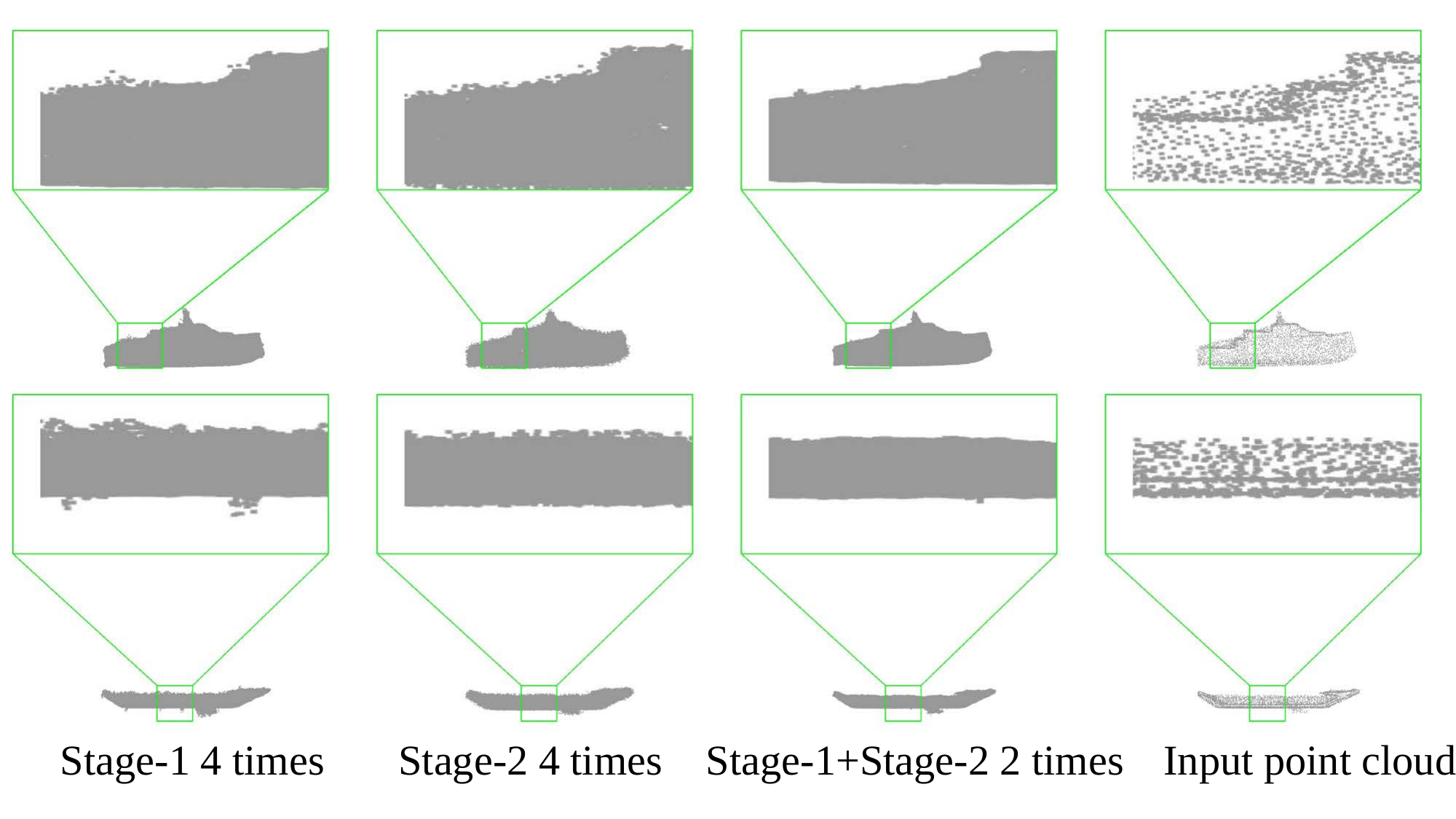}
  \caption{The visualization result of 2-stage UDF prediction ablation study. It is 16$\times$ upsampled-point cloud comparison.}
  \label{fig:stage}
\end{figure}

%% file: figures/normal.tex
\begin{figure}
\centering
  \includegraphics[width=\columnwidth]{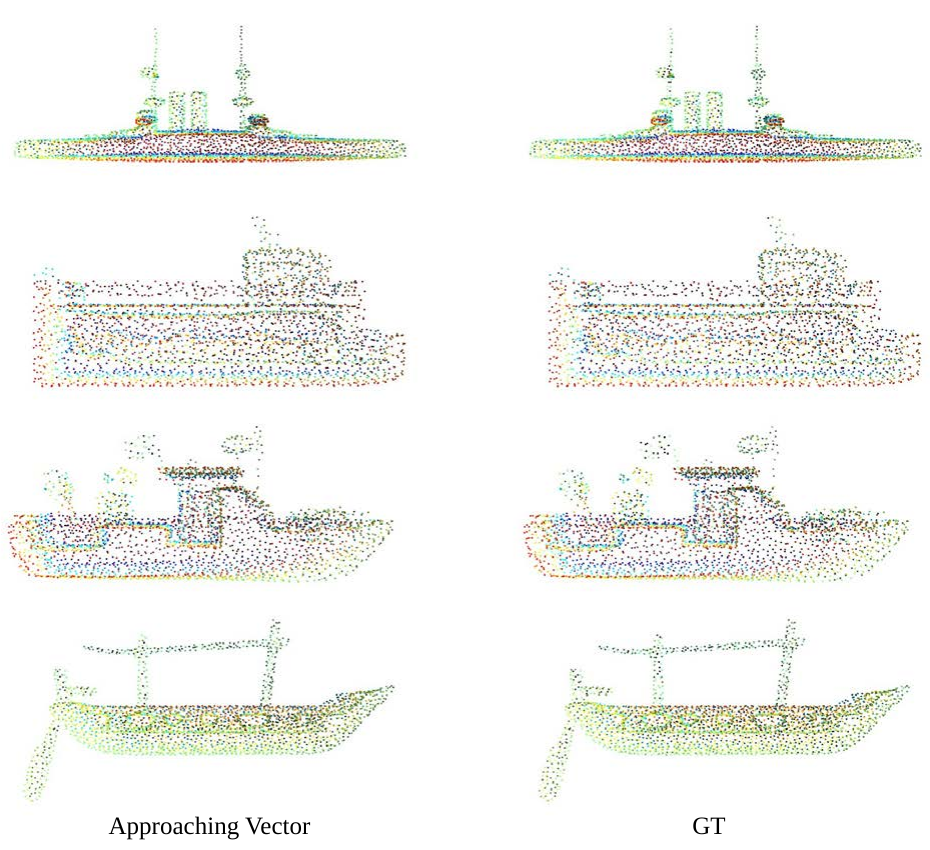}
  \caption{The visualization result of approaching vector. The color is mapped from approaching vector. We can see that the approaching vector is close to the ground-truth normal.}
  \label{fig:normal}
\end{figure}

%% file: figures/compare_meshudf.tex
\begin{figure}
\centering
  \includegraphics[width=\columnwidth]{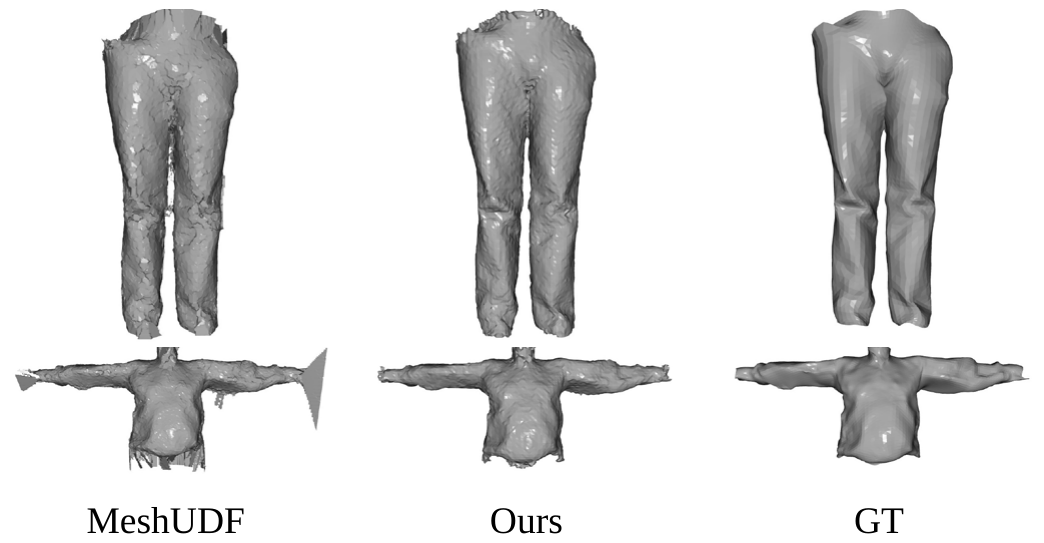}
  \caption{The visualization of comparison between our mesh extraction method and MeshUDF on predicted UDF.}
  \label{fig:compare_meshudf}
\end{figure}

%% file: figures/famous_result.tex
\begin{figure} 
\centering
  \includegraphics[width=\columnwidth]{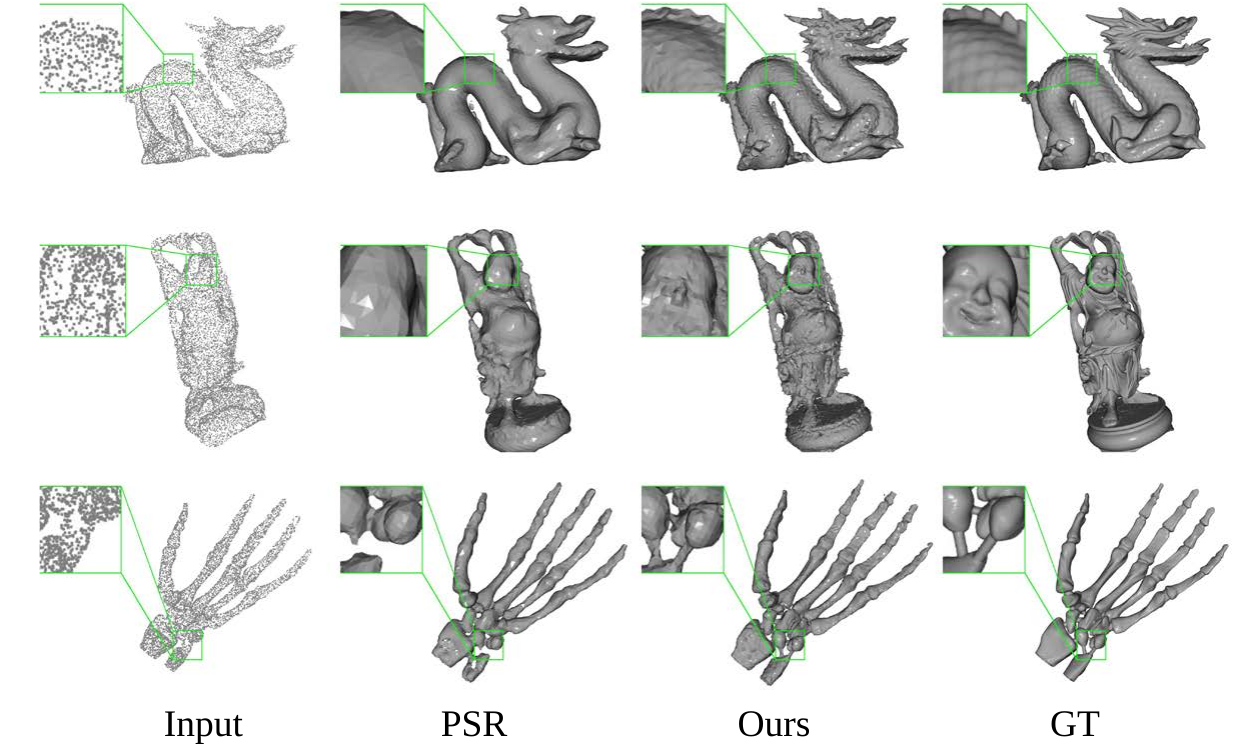}
  \caption{Visualized results of our method and PSR\cite{2013Poisson} on FAMOUS dataset.}
  \label{fig:famous_result}
\end{figure}

%% file: 5conclusion.tex
\section{Conclusion and Future Work}
In this work, we propose the self-supervised pipeline to reconstruct the implicit surface based on UDF with a regularization term. Through this pipeline, we can reconstruct good UDF field. In order to reconstruct mesh directly from UDF, we propose a learning-based iso-surface extraction and achieve good result. However, one weakness is the edge of the open surface, our method usually generates some noise. Because the UDF distribution on the open surface edge is hard to handle. Our next work will focus on improving reconstruction quality of the surface edge.
\section{Acknowledgements}
This research is funded by the following projects. National Key Research and Development Program of China(2018AAA0102200). National Natural Science Foundation of China(62002375, 62002376, 62132021, 62325211, 62372457, 62002379). National Science Foundation of Hunan Province of China(2021JJ40696, 2021RC3071, 2022RC1104, 2023JJ20051). NUDT Research Grants(ZK22-52). The Science and Technology Innovation Program of Hunan Province(2023RC3011).

%% file: 6appendix.tex
\appendix
\input{figures/motivation.tex}
\subsection{The Motivation of self-supervised UDF prediction}
\label{motivation}

Currently, almost all the learning methods are supervised methods. However the 3D ground truth is very expensive to acquire. Further more, the experiment can only be done on synthesis dataset, such as ShapeNet~\cite{chang2015shapenet}, because real world dataset usually do not contain perfect ground truth surface. To apply implicit surface reconstruction from point cloud to real world dataset. One way~\cite{2020SSRNet} is training the network on synthesis dataset and generalize the method to real world dataset. However, this idea is also restricted when point cloud density and local shape is much dissimilar to the training data, the performance will degrade as well. Another idea is training the implicit surface reconstruction network in an unsupervised manner. Then, the method will not be disturbed by above problems. Because we can train the network on target dataset even if the target dataset do not have 3D ground truth surface. 
When we want to modify the original supervised pipeline to unsupervised one, in Fig. \ref{fig:motivation} (a), we have to overcome two problem. Marching Cube and point cloud sampling is not differentiable. Although we can follow the method of deep marching cube~\cite{2018Deepmc}, we cannot find a light differentiable sampling method on mesh. However, if we consider the two non-differentiable methods together, we can design a method directly sampling point cloud on UDF implicit surface. That is shifting the query point along the UDF gradient direction and forwarding the distance of UDF value. Then the query point will reach the zero-level set of the UDF field. The pipeline is shown in Fig. \ref{fig:motivation} (b). Thus, we can train the network with chamfer loss.

\subsection{Evaluation Metric}

We also provide the quantitative result with modified normal consistency as in Eq. (\ref{eqn:nc}),
\begin{equation}
\label{eqn:nc}
\begin{aligned}
\mathrm{NC}=& \frac{1}{2 N_{x}} \sum_{i=1}^{N_{x}}|<\mathbf{n}\left(\mathbf{x}_{i}) , \mathbf{n}(\mathcal{S}_{y}\left(\mathbf{x}_{i}\right)\right)>|+\\
& \frac{1}{2 N_{y}} \sum_{i=1}^{N_{y}}|<\mathbf{n}\left(\mathbf{y}_{i}) , \mathbf{n}(\mathcal{S}_{x}\left(\mathbf{y}_{i}\right)\right)>|.
\end{aligned}
\end{equation}
where $\{ \mathbf{x}_{i}, i=0\cdots N_x\}$ is the points set sampled from ground-truth mesh. $\{\mathbf{y}_{i}, i=0\cdots N_y\}$ is the point set sampled from reconstructed mesh. $\mathcal{S}_{y}(\mathbf{x}_{i})$ means the nearest point to $\mathbf{x}_{i}$ in reconstructed point set. $\mathcal{S}_{x}(\mathbf{y}_{i})$ means the nearest point to $\mathbf{y}_{i}$ in ground-truth point set. $|\cdot|$ means absolute value. $\mathbf{n}(x)$ means the normal of point $x$. $<,>$ means inner-product.

\begin{table}[!t]
  \centering
  \scalebox{0.75}{
  \begin{tabular}{l|lll}
    \toprule
    Methods & Field prediction & Mesh extraction & Total\\
    \midrule
    NPull~\cite{2020npull} & 29min22s & \bf{1.2s} & 29min23s\\
    CAP~\cite{capudf} & 21min56s & 15s & 22min11s\\
    Ours & \bf{9s} & 46s & \bf{55s}\\
    \bottomrule
  \end{tabular}
  }
  \vspace{3pt}
  \caption{Inference efficency comparison.}
      \label{tab:efficiency}
\end{table}

\subsection{Effectiveness Comparison}
CAP~\cite{capudf} and Neural Pull~\cite{2020npull} directly predict the distance function with coordinate of query point as input and point cloud is only used as loss function during training, thus they need training the network even in test time. Ours first calculate point-wise feature, then for every query point, we calculate its feature and predict the distance function. Thus, ours do not need train the network during test time, that's the reason why ours is faster than CAP-UDF and Neural Pull in distance prediction. However, during mesh extraction stage, because Neural Pull is based on SDF, it can extract mesh directly making use of the Marching Cube. CAP-UDF can predict the relative sign with a simple hand-designed rule. Thus, they are fast than ours in this stage. 

The result is shown in Tab. \ref{tab:efficiency}. What's more, in the mesh extraction stage, there are lots of for-loop structure in the code. And our code is implemented with python and without any parallel program. While each cube can be addressed separately. Thus the execution time of ours and CAP-UDF can be shorten by writing C code with openmp.

\subsection{Slice Visualization}
In order to show what the network indeed learn, we visualize the direct output of the network, UDF slices along x, y and z axis. 
We visualize the UDF in 2D slices in Fig. \ref{fig:slice}. We can see that in all the 3 slices from x-axis, y-aixs or z-aixs, the UDF is close to 0 near the surface, gradually increases when the distance to surface is larger. The relative error between our prediction and the ground truth is only 3\% which demonstrate that the network indeed learns correct UDF distribution around the surface. 

\input{figures/slice.tex}

\subsection{Learning-based Iso-surface Extraction}
\subsubsection{The Motivation of Learning based Iso-surface Extraction}

First, we introduce a key observation of mesh reconstruction. Let's take a look at Marching Cube. It is a template match method. For every cube, according to the sign of the eight corners, it can generate a corresponding mesh in the cube. However, if we flip the sign of the eight corners at the same time, the only change of the generated mesh is that the normal of the triangulation is flipped. In most occasions, we do not need global consistent triangulation orientation if our aim is to recover the shape of the implicit surface. Thus, we can make a minor sacrifice of the reconstructed mesh, giving up the globally consistent triangulation orientation, to make the sign prediction much easier. 

Next, we introduce the evolution of our learning based iso-surface extraction idea. Inspired by ~\cite{ioffe2015batch}, we want to expand their 1D quadratic function on the edge to 3D quadratic function in the space to overcome the sign conflict problem in ~\cite{ioffe2015batch}. However, due to the noise, it is not a very stable solution. 

Thus, we want to treat the sign prediction problem as a classification problem by giving up the curve as a middle result. Luckily, the UDF has already been voxelized and can be easily processed by standard 3D convolution network. Therefore, we just make use of 3D convolution to extract the feature of 8 corner vertices in the cube. We can not only make use of the function value, but also make use of the partial derivatives to make the curve fitting more reliable. Thus, inspired by the curve fitting process, the network input not only include the UDF value, but also the gradient. Further more, we can add more information by adding sample points in the cube to improve the input feature quality. Thus, our final input for network is a $4 \times 5 \times 5 \times 5$ cube. The first channel include 1-dimension UDF value and 3-dimension UDF gradient.

If we adopt a learning method, we can treat the sign predict problem as classification problem. Considering the key observation mentioned above, we have to handle the relative sign prediction problem in learning method pipeline. In detail, if the network predicts the eight corner's sign which is the same with the ground-truth or is the flip of the ground-truth sign, the loss should be 0. In order to handle this problem, we design 3 methods, all of them can achieve rather good result. 

\subsubsection{Alternatives of Sign Prediction in Learning Based Iso-surface Extraction}
In learning based iso-surface extraction part, we need to handle the relative sign problem. We also propose 3 ways to handle it. First, we use a modified binary cross entropy loss to predict the relative sign of the UDF in the corner of cube. In detail, we adopt the following 
\begin{align}
    L_1 = \sum_{c \in \mathcal{C}}min(\sum_{v \in V}BC(\Tilde{y}^{c}_j, y^{c}_j), \sum_{j=0}^{7}BC(\Tilde{y}^{c}_j, \neg y^{c}_j)),
\end{align}
where $\mathcal{C}$ is the cube set, $c$ is the cube index, $j$ is vertex index in the local cube. $BC$ is binary cross entropy, $\Tilde{y}$ is the predicted sign, $y$ is the ground-truth label, $y \in \{0, 1\}$, $\neg y$ is the negative $y$. In this way, we choose a label or negative label which is close to the prediction. In testing, we only need to obtain one prediction same with label or negative label. Obviously, from the UDF and its gradient in the local cube, it is impossible to infer the global sign. On the contrary, predicting relative sign is possible.

In fact, predicting the relative sign of vertex is learning the relation between vertices. Thus, we can give up the above way where we treat the eight vertices as a whole. Every time, we only handle two vertices and predict their relation. We can write the loss as 
\begin{align}
    L_2 = \sum_{c \in \mathcal{C}}\sum_{i,j=0}^{7}BC(\Tilde{y}^{c}_{ij}, y_i \oplus y_j),
\end{align}
where $\Tilde{y}_{ij}$ is the relation of $i-th$ vertex and $j-th$ vertex. $\oplus$ is xor. 

Thirdly, we can go further from the first idea, treating the cube as a whole. we do not predict the sign the each vertex but predict the whole status of the cube. In this way, we can totally have $128$ cube status. Thus, the sign prediction problem becomes a multi-classification problem. The loss is 
\begin{align}
    L_3 = \sum_{c \in \mathcal{C}}CE(\Tilde{y^c}, y^{c})
\end{align}
where $y \in \{0, 1, \cdots, 127\}$, $CE$ is cross-entropy. It shall be explained here why the number of status is 128. Totally, we have eight vertex and varies in $\{0, 1\}$, therefore, there are $2^8=256$ status. Because we regard the flip of all the vertices and the original status as the same, thus the number of status is $256/2=128$.

We define the acc here. For every the cube near the surface, $c_i, i=1,..,N$, if the network predicts the right sign of all the 8 corners or the negative sign of all the 8 corners, the indicator function $\mathbb{I}(c_i)=1$, otherwise, $\mathbb{I}(c_i)=0$. Then the metric is the accuracy defined as 
 \begin{equation}
     acc = \sum_i^N\frac{\mathbb{I}(c_i)}{N}.
     \label{eq:acc}
 \end{equation}
 We compare the three methods for predict the cube status and the results do not vary too much. The comparison result is shown in Tab. \ref{tab:loss}. We random choose 20 shapes from the all the 3 datasets for training.
\begin{table}[]
\centering
\begin{tabular}{llll}
\toprule
Loss     & $L_1$  & $L_2$  & $L_3$  \\
\midrule 
ShapeNet & 0.97 & 0.96 & 0.97 \\
Garment  & 0.97 & 0.97 & 0.96 \\
ScanNet  & 0.96 & 0.96 & 0.95 \\
\bottomrule
\end{tabular}
\vspace{3pt}
\caption{The acc(\ref{eq:acc}) result training with different loss.}
\label{tab:loss}
\end{table}

\subsubsection{Sign Conflict}
The term sign conflict has two different meanings in different situations. One situation is that the two corners of two consecutive cubes have different sign by prediction in SDF reconstruction, but the two corners are the same point in fact, they should have the same sign. Another is in UDF reconstruction, the sign of 8 corners is conflict with each other. For simplicity, taking a sqaure rather than cube for example, as shown in Fig. \ref{fig:topology}. A and B have opposite UDF gradient so they have the opposite sign. Let's suppose A is positive and B is negative. Because A and C have the opposite sign as well, there C is negative. C and D have the same-direction gradient, therefore they have the same sign, D is negative. While, B and D have opposite gradient, they have different sign, D is positive. D is positive and negative at the same time, there is conflict. When conflict arises, we cannot map the sign of cube corner to the Marching Cube table, therefore we cannot generate mesh as well. Here, the sign conflict means the latter one.
The mesh extraction method of CAP~\cite{capudf} has a risk of sign conflict. There are 8 corners of every cube, so there are $2^8=256$ possible situations of the corner sign composition. Meanwhile if we flip the sign of 8 corners at the same time, the mesh without orientation does not change. Therefore, there are totally $256/2=128$ corner sign composition. While it extract mesh via predicting whether the two consecutive corners have the same sign separately. There are 12 edges of the cube so that there are $2^{12}=2048$ sign composition. There are $2048-128 = 1920$ edge sign composition generating sign conflict. 
\input{figures/topology}
However, if adopting our method to reconstruct mesh, we directly output the sign of the 8 corners by considering the cube as a whole rather than considering the pair of corners separately. Although the sign conflict does not always appear in practice, our method for mesh extraction is complete in theory.

\subsubsection{Details of Learning based Iso-surface Extraction}
In order to improve the generalization ability of our network further, We design three ways to augment the input data. First, we adopt random flip. In every training iteration, we randomly choose a axis and give a probability of whether flip or not. Second, we adopt random rotation. we randomly choose a axis and give a probability of rotation angle $0^{\circ}, 90^{\circ}, 180^{\circ}, 270^{\circ}$. Third, we adopt random scale. We randomly scale the UDF value in the cube, the scale parameter varies in $[0.5, 1.5]$.

The network is backbone is composed of 4 3D convolution layers. The width of the layers is $8$, $64$, $128$ and $128$. After extracting the feature, we use a 3D positional encoding~\cite{2020Fourier} to encode the coordinates of 8 corners, then concatenating the feature of the cube and 8 corners. During training, we adopt cosine learning rate adjustment~\cite{2016coslr}, the initial learning rate is $0.001$. We use Adam~\cite{adam} optimizer to learn the parameter. The batch-size is $2048$.

 In our experiment, we train the network on ShapeNet vessel class with different training shapes and test on other classes of ShapeNet. The result is shown in Tab. \ref{tab:cross_class}. We can conclude that only 5 shapes for training is good enough and the network can generalize to different classes.
\begin{table}[]
\begin{tabular}{lllll}
\toprule
Test Class         & airplane & bench & cabinet & rifle \\
\midrule
5 training shapes  & 0.95     & 0.96  & 0.96    & 0.97  \\
10 training shapes & 0.95     & 0.96  & 0.96    & 0.96  \\
20 training shapes & 0.96     & 0.97  & 0.96    & 0.97  \\
50 training shapes & 0.96     & 0.97  & 0.96    & 0.96 \\
\bottomrule
\end{tabular}
\vspace{3pt}
\caption{The accuracy training on vessel of ShapeNet and test on different classes.}
\label{tab:cross_class}
\end{table}

%% file: figures/motivation.tex
\begin{figure*} 
\centering
  \includegraphics[width=2\columnwidth]{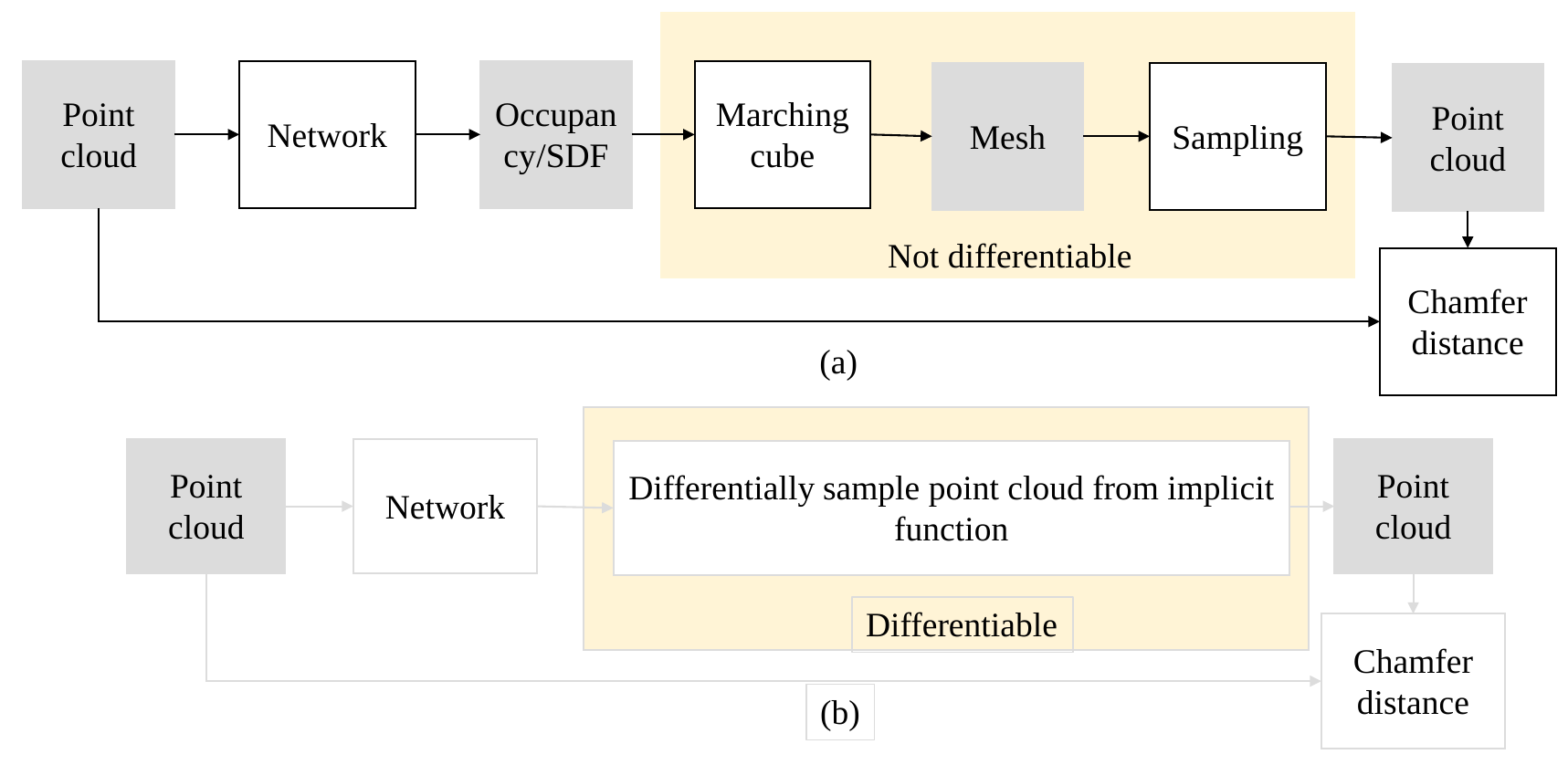}
  \caption{Motivation of UDF prediction method. (a) is pipeline based in SDF. (b) is our pipeline.}
  \label{fig:motivation}
\end{figure*}

%% file: figures/slice.tex
\begin{figure}
\centering
  \includegraphics[width=\columnwidth]{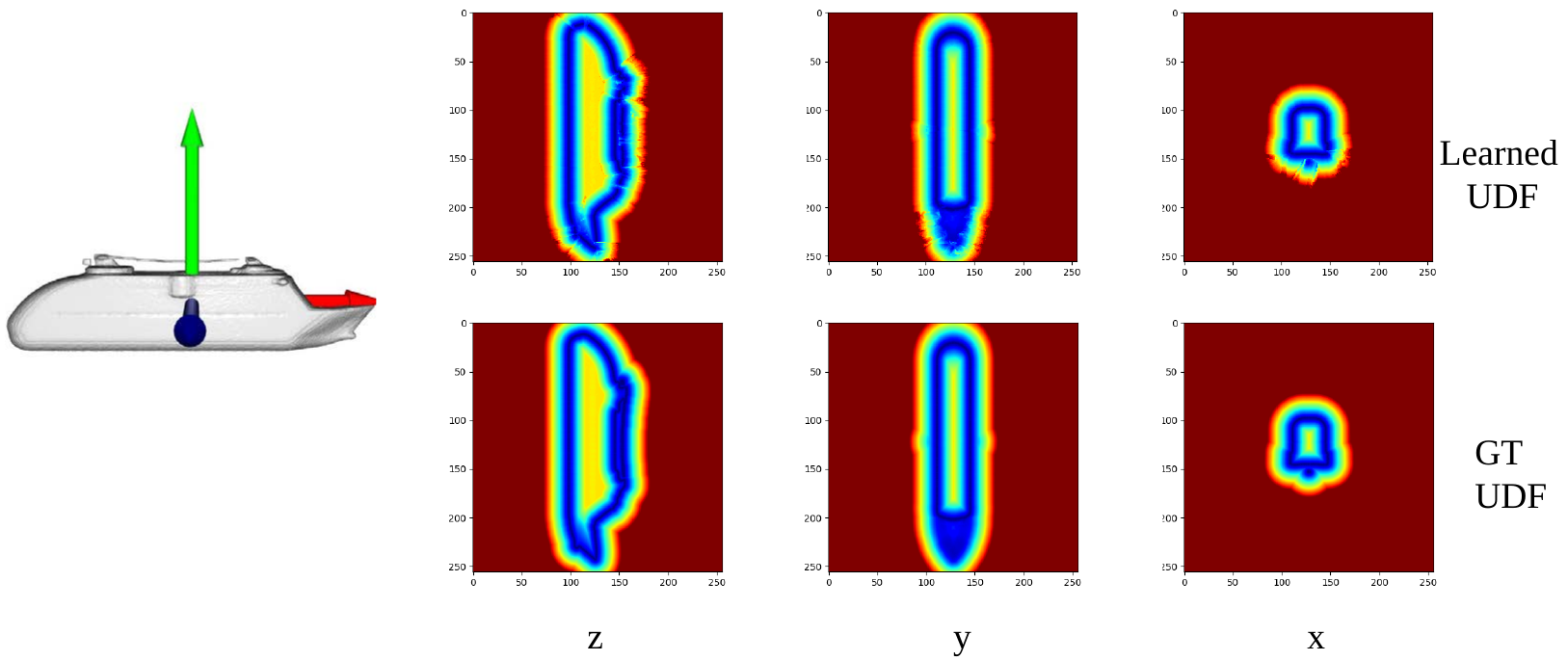}
  \caption{The 2D UDF value slices along x, y and z axis. The color red is large and blue is small.}
  \label{fig:slice}
\end{figure}

%% file: figures/topology.tex
\begin{figure} 
\centering
\includegraphics[width=0.5\columnwidth]{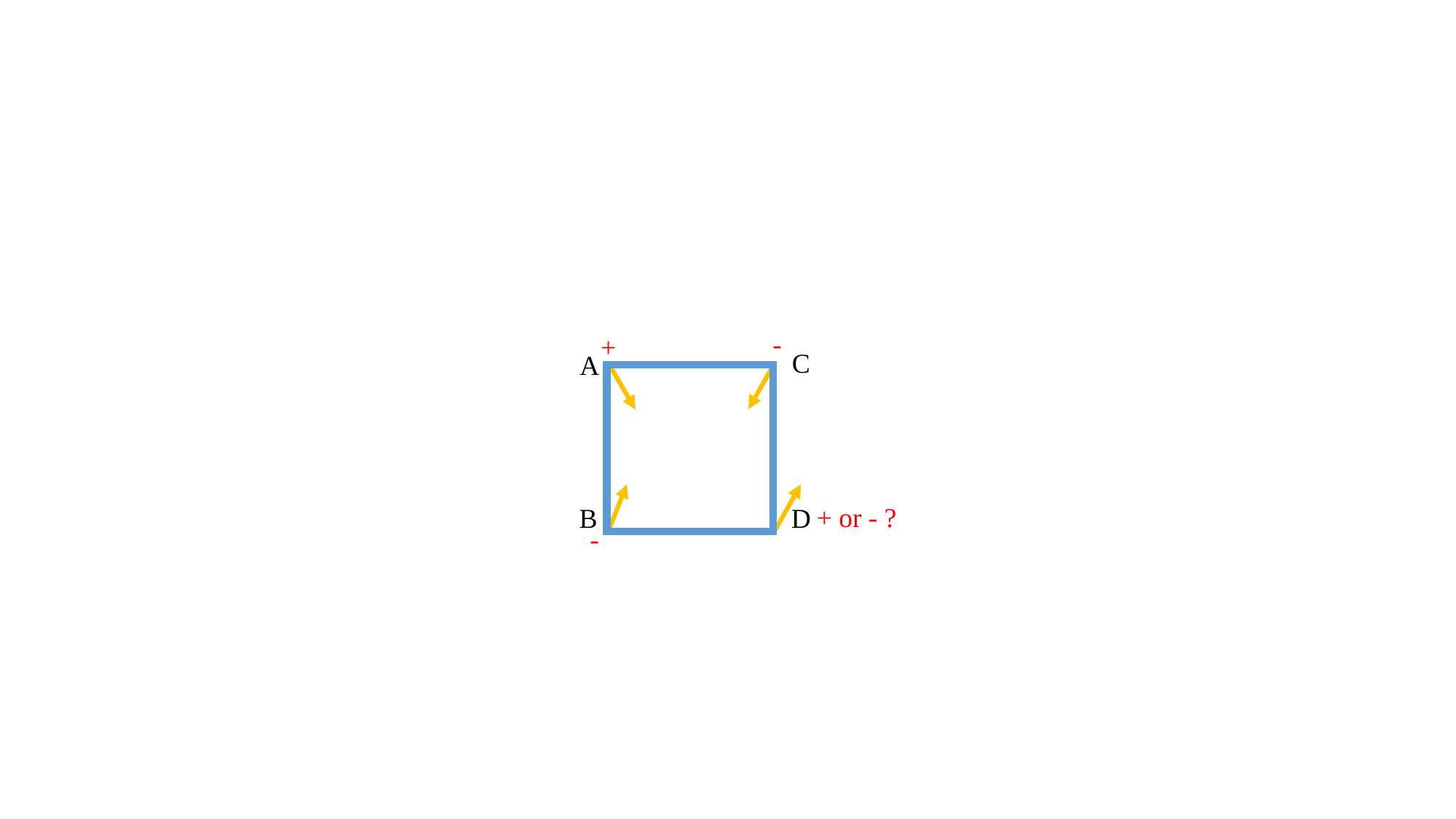}
  \caption{The sign conflict of mesh extraction. The orange arrow is the UDF gradient of every point.}
  \label{fig:topology}
\end{figure}